\documentclass[journal]{IEEEtran}
\usepackage{graphicx}
\usepackage{graphics}
\usepackage{epsfig}
\usepackage{epstopdf}
\usepackage{stfloats}

\usepackage{xcolor,soul,framed} 

\colorlet{shadecolor}{yellow}
\usepackage[cmex10]{amsmath}
\usepackage{array}
\usepackage{mdwmath}
\usepackage{mdwtab}
\usepackage{eqparbox}
\usepackage{url}
\usepackage[colorlinks,
            linkcolor=red, 
             anchorcolor=blue, 
            citecolor=green, 
            ]{hyperref}
\usepackage{float}
\usepackage[ruled,linesnumbered,vlined]{algorithm2e}
\usepackage{algorithmic}
\usepackage{mathrsfs}
\usepackage{amssymb}
\usepackage{amsthm}
\usepackage{cite}
\usepackage{url}
\usepackage{xcolor}
\usepackage{subfig}
\usepackage{fancyhdr}
\usepackage{mdwtab}
\usepackage{caption}
\usepackage{stfloats}
\usepackage{array}
\usepackage{bbding}
\usepackage{subfloat}
\usepackage{verbatim}
\usepackage{threeparttable}

\usepackage[normalem]{ulem}
\usepackage{makecell}

\begin{document}
\bstctlcite{IEEEexample:BSTcontrol}
    \title{A Flying Bird Object Detection Method for Surveillance Video}
  \author{Zi-Wei Sun,
          Ze-Xi Hua,
      Heng-Chao~Li,~\IEEEmembership{Senior Member,~IEEE}, and Yan Li
  \thanks{Zi-Wei Sun, Ze-Xi Hua, Heng-Chao Li, and Yan Li are with the School of Information Science and Technology, Southwest Jiaotong University, Chengdu 611756, China (e-maile: xx\_zxhua@swjtu.edu.cn).}
  }

\markboth{Arxiv}{Ziwei \MakeLowercase{\textit{et al.}}: A Flying Bird Object Detection Method for Surveillance Video}

\maketitle

\begin{abstract}
Aiming at the specific characteristics of flying bird objects in surveillance video, such as the typically non-obvious features in single-frame images, small size in most instances, and asymmetric shapes, this paper proposes a Flying Bird Object Detection method for Surveillance Video (FBOD-SV). Firstly, a new feature aggregation module, the Correlation Attention Feature Aggregation (Co-Attention-FA) module, is designed to aggregate the features of the flying bird object according to the bird object's correlation on multiple consecutive frames of images. Secondly, a Flying Bird Object Detection Network (FBOD-Net) with down-sampling followed by up-sampling is designed, which utilizes a large feature layer that fuses fine spatial information and large receptive field information to detect special multi-scale (mostly small-scale) bird objects. Finally, the SimOTA dynamic label allocation method is applied to One-Category object detection, and the SimOTA-OC dynamic label strategy is proposed to solve the difficult problem of label allocation caused by irregular flying bird objects. In this paper, the performance of the FBOD-SV is validated using experimental datasets of flying bird objects in traction substation surveillance videos. The experimental results show that the FBOD-SV effectively improves the detection performance of flying bird objects in surveillance video. This project is publicly available \href{https://github.com/Ziwei89/FBOD}{https://github.com/Ziwei89/FBOD}.
\end{abstract}


%
\IEEEpeerreviewmaketitle


\section{Introduction}

\IEEEPARstart{B}{IRD} detection plays an extremely important role in ensuring safety and alleviating conflicts between humans and animals and has attracted more and more researchers’ attention in recent years. At present, the method of detecting bird objects by radar is widely used \cite{2016_Hoffmann_multistatic_radar, 2017_Jahangirstaring_radar}, but the radar equipment has the disadvantages of being large in volume,  expensive, and having poor visual effects. With the development of computer vision, deep learning, and other technologies, there are more and more studies on using cameras to detect various objects \cite{2022_Ye_CT-Net, 2022_Zheng_Foreign_Objects, 2023_Ni_Improved_SSD, 2023_Ye_Real-Time_Object_Detection_Network_in_UAV-Vision}. We are working on the real-time detection of flying birds using surveillance cameras. However, there are three main challenges for detecting flying birds in surveillance video.

\begin{figure*}[htb]
\centering
\subfloat[The characteristics of the flying bird are not obvious]{\includegraphics[width=6.5in]{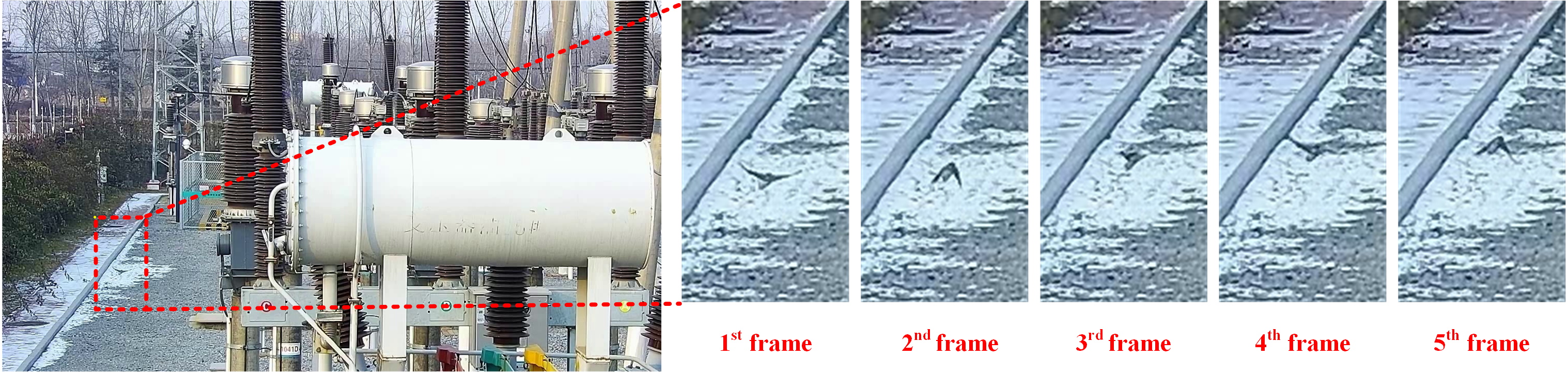}
\label{obvious_bird_fig}}
\\
\subfloat[The flying bird has a special multi-scale property]{\includegraphics[width=2.45in]{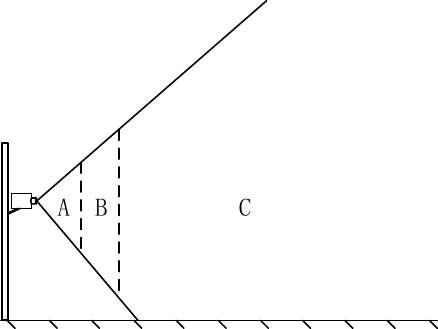}
\label{special_scale_bird_fig}}
\subfloat[The flying bird in the bounding box is irregular]{\includegraphics[width=2.8in]{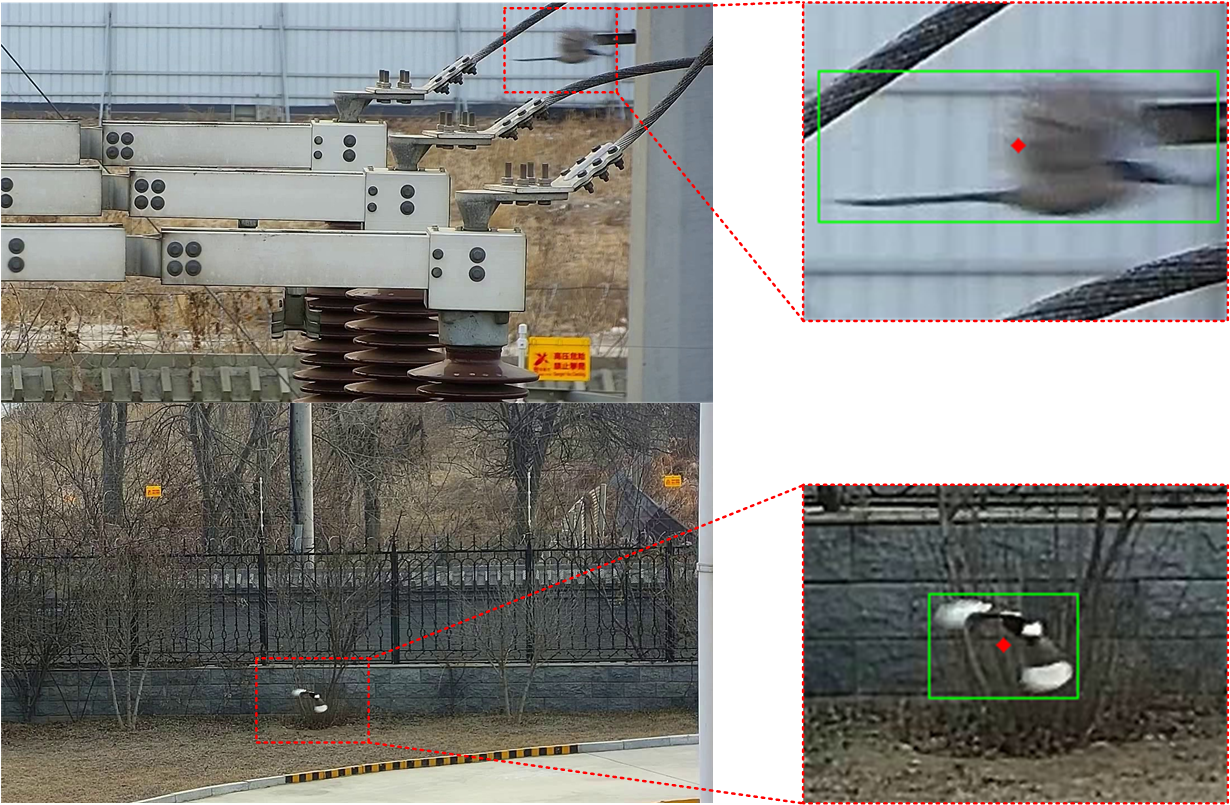}
\label{irregularity_scale_bird_fig}}
\caption{Characteristics of flying bird objects in surveillance videos. (a) On the right, there is a small bird with weak features. Left is a screenshot of the bird on five consecutive frames. (b) Simplified diagram of the distribution of flying birds in surveillance video. Birds are evenly distributed in the surveillance area (A, B, and C areas: birds in area A are large objects, birds in area B are general objects, and birds in area C are small objects). However, the space of area C is much larger than that of areas A and B, so birds are mostly small objects. (c) In the right image, the green box is the bounding box, and the red point is the middle of the bounding box. Due to the shape of the bird object itself, it is not regular in the bounding box in most cases.}
\label{Flying_bird_in_surveillance_videos}
\end{figure*}

\begin{enumerate}
\item{The characteristics of the single frame image of the flying bird object in the surveillance video are not obvious. In most cases, the flying bird object blends into the background, making it challenging to distinguish from the environment, even after careful examination, as shown on the left in Fig. \ref{Flying_bird_in_surveillance_videos}\subref{obvious_bird_fig}.}
\item{Most flying bird objects in the surveillance video are small.  As Fig. \ref{Flying_bird_in_surveillance_videos}\subref{special_scale_bird_fig} demonstrates, when birds are far from the camera, they occupy fewer pixels in the video frame, categorizing them as small objects.  Conversely, when birds are closer to the camera, they occupy more pixels and can be considered large objects.  However, birds are often far away from the camera, resulting in them being predominantly classified as small objects.}
\item{In most cases, the flying birds in the surveillance video are not symmetrical. For example, in some cases, the body and tail of the bird are divided roughly in half within the bounding box, with half of the tail occupying more background pixels. When the flying bird spreads its wings, the part without wings contains more background pixels, as shown in Fig. \ref{Flying_bird_in_surveillance_videos}\subref{irregularity_scale_bird_fig}.}
\end{enumerate} 

Currently, most object detection algorithms based on computer vision \cite{2014_Girshick_RCNN, 2015_Girshick_Fast_RCNN, 2017_Ren_Faster_RCNN, 2016_Liu_SSD, 2016_Redmon_YOLO, 2017_Redmon_YOLOV2, 2018_Redmon_YOLOv3, 2020_Bochkovskiy_YOLOv4, yolov5_2021, 2021_Zheng_YOLOX, 2022_Chuyi_yolov6, 2023_Wang_yolov7, 2023_Guang_yolov8} do not specifically conduct in-depth research on the flying bird object in the actual surveillance video but treat it as an ordinary object. This class of methods achieves surprising results on generic objects by extracting features of the object and then performing classification and regression. However, the effect is unsatisfactory when applied to the flying birds in the surveillance video. The main reason for this phenomenon is that the flying bird objects in the datasets they rely on have obvious features and clear appearances, such as the flying bird objects in ImageNet \cite{2015_ImageNet}, COCO \cite{2014_COCO}, PASCAL VOC \cite{2010_Pascal_VOC}, and other datasets.

Some papers have specifically studied the detection methods of flying bird objects.  For example, literature \cite{2018_Tianhuang_skeleton_flying_bird} proposed a skeleton-based bird object detection method, which describes the motion information of birds through a set of key postures. However, this method uses the moving object detection method based on VIBE \cite{2011_Barnich_ViBe} when extracting the contour of the flying bird object. Therefore, this method is unsuitable for situations where the flying bird object is temporarily stationary or the background dynamic changes sharply, which is common in surveillance video.

Another example is Glances and Stare Detection(GSD) \cite{2019_Tian_GSD}, which simulates the characteristics of human observation of small moving objects. Firstly, the whole image is glanced at, and the places containing objects are carefully observed. Essentially, this method still relies on single-frame image features (see Algorithm 1 in reference \cite{2019_Tian_GSD}), thus requiring that the single-frame image features of the flying bird object are clear enough. However, the task in our paper needs to deal with the flying bird objects with unobvious features in single frame images, and the method based on single frame image features cannot achieve good results because the features of these objects are easy to lose in the process of feature extraction (see the comparative experiment in Section \ref{ablation_analysis_1}).

In this paper, we consider many characteristics of flying birds in surveillance video (single image features are not obvious characteristics, special multi-scale properties, irregular properties in the bounding box) and propose a suitable detection method for flying birds in surveillance video, FBOD-SV.

Aiming at the problem that the features of the single frame image of the flying bird object in the surveillance video are not obvious, our previous work \cite{2024_sun_Flying_Bird} used the Convolutional Long Short-Term Memory network (ConvLSTM) to aggregate the spatio-temporal information of adjacent multiple frames of flying bird objects at the input of the model to enhance the characteristics of flying bird objects, to improve the detection rate of flying bird objects. However, ConvLSTM suffers from slow runtime and inadequate information aggregation.

Although the characteristics of the single frame image of the flying bird object in the surveillance video are not obvious in some cases, the object can still be found by observing the continuous multiple frames of images, as shown on the right in Fig. \ref{Flying_bird_in_surveillance_videos}\subref{obvious_bird_fig}. Therefore, in some cases, the information from other image frames can be used to observe the object on one frame because the bird object has a correlated relationship between frames. In this paper, we take advantage of this correlation relationship and propose a new method to aggregate the information of consecutive multiple frames of images, the Correlation Attention Feature Aggregation (Co-Attention-FA) method, which uses the correlation relationship of flying bird objects on consecutive multiple frames of images to ignore redundant background information and focus on the region containing the flying bird object.

Aiming at the special multi-scale property of flying birds in surveillance videos (most of them belong to small objects), a Flying Bird Object Detection Network (FBOD-Net) is designed, which performs down-sampling followed by up-sampling. A large feature layer that combines fine spatial information and large receptive field information is used to detect flying birds in surveillance videos. The large feature layer has certain advantages for detecting small-scale size bird objects because it has fine spatial information (shallow, high-resolution feature layer) and highly abstract semantic information (deep feature layer). Due to the fusion of large receptive field information, the large feature layer also has the ability to detect large-scale size flying bird objects \footnote{FBOD-Net is trained on datasets dominated by small objects with occasional large objects and can detect large objects. However, the training process may not converge if datasets contain mainly large objects.}.

Aiming at the irregular characteristics of the bird object in the bounding box in the surveillance video, the SimOTA dynamic label strategy for One-Category object detection (SimOTA-OC) is proposed. When the SimOTA \cite{2021_Zheng_YOLOX} dynamic label allocation method is applied to the bird object detection (one-category object detection), only the IOU between the Ground Truth (GT) box and the predicted box is used to realize the dynamic allocation of labels, which solves the problem of inaccurate label allocation caused by the irregular bird object.

The main contributions of this paper are as follows.

\begin{enumerate}
\item{The Correlation Attention Feature Aggregation (Co-Attention-FA) method is proposed to make use of the correlation of bird objects on consecutive frames so that the model can pay attention to the bird objects with unobvious features on a single frame image. This method improves the detection rate of flying birds and reduces the time consumed by information aggregation of consecutive multiple frames of images.}
\item{Aiming at the special multi-scale property of flying birds in surveillance video, which belongs to small objects in most cases, a model structure, Flying Bird Object Detection Network (FBOD-Net), is designed to predict flying birds using only one large feature layer. The large feature layer fully integrates the information of deep and shallow feature layers and can classify and locate the flying birds in the surveillance video well.}
\item{The SimoTA-OC dynamic label allocation method is proposed. For One-Category object detection, when implementing the SimOTA dynamic label allocation method, only the IOU between the GT box and the predicted box is used to realize the dynamic allocation of labels, which solves the problem of inaccurate label allocation caused by the irregular shape of the flying bird object.}
\end{enumerate}

The remainder of this paper is structured as follows: Section \ref{Related Work} presents work related to this paper. Section \ref{The Proposed Methord} describes the proposed method of detecting flying bird objects in surveillance video in detail. In Section \ref{Experiment}, the ablation and comparison experiments of the proposed algorithm (method) are carried out. Section \ref{Conclusion} concludes our work.

\section{Related Work}\label{Related Work}

In this paper, we mainly study the three characteristics of the flying bird object in surveillance video (the characteristics of the single frame image are not obvious, the special multi-scale attribute, and the irregular attribute in the bounding box), which mainly involves the feature aggregation, multi-scale feature extraction and label assignment related work. Therefore, this part mainly introduces the related work of feature aggregation, multi-scale object detection, and label assignment.

\subsection{Related Work on Feature Aggregation}\label{Feature_aggregation}

The related research work of feature aggregation mainly deals with video object detection. When the features of the object to be detected on some video frames are not obvious (appearance change, occlusion, motion blur), the feature information of other adjacent frames can be aggregated to enhance the features of the object on these frames in the feature extraction stage. For example, Zhu Xizhou et al. \cite{2017_Zhu_DFF, 2017_Zhu_FGFA, 2018_Zhu_Towards_High_Performance, 2017_Hetang_Impression_Network} used optical flow to propagate features, which were then aggregated with the features of the current frame to enhance the features of the object. Haiping Wu et al. \cite{2019_Wu_SELSA} proposed a SEquent-Level-Semantic-Aggregation (SELSA) method to enhance the features of the object in the candidate box. Tao Gong et al. \cite{2021_Tao_Temporal_RoI_Align} aggregated the ROI features of the current frame and the most similar ROI features of other frames to obtain the Temporal ROI features of the object.

In the previous work \cite{2024_sun_Flying_Bird}, we pointed out that the above method of feature aggregation, whose aggregation operation is carried out in the intermediate feature layer, is unsuitable for the flying bird object in the surveillance video. Because the appearance features of most of the bird objects in the surveillance video are not particularly rich in any single frame, and the object size is small, the features of the object are easy to lose when extracting the features of a single frame image. After feature extraction, aggregation is easy to introduce some wrong information. Therefore, in the previous work \cite{2024_sun_Flying_Bird}, we used the ConvLSTM to aggregate the spatio-temporal information of adjacent multiple frames of flying bird objects at the input of the model to enhance the characteristics of flying bird objects, to improve the detection rate of the flying bird object. However, ConvLSTM suffers from slow runtime and inadequate information aggregation.

In this paper, we propose a new method to aggregate the information of consecutive multiple frames of images, the Co-Attention-FA method, which takes advantage of the correlation between the bird objects in consecutive multiple frames of images, ignores the redundant background information and focuses on the region containing the bird object.

\subsection{Related Work on Multi-scale Object Detection}\label{multi-scale_object}

Multi-scale detectors need to be designed to detect objects of different sizes (different scales). In the beginning, to deal with the multi-scale attribute of the object, the scheme adopted is to build an image pyramid and use the image at the bottom of the pyramid to detect small objects and the image at the top of the pyramid to detect large objects (that is, use images at different scales to detect objects at different scales). Both traditional object detection methods and early deep learning-based object detection methods adopt this approach when dealing with multi-scale problems. However, this method needs to extract features from images of multiple scales, which consumes a lot of time and computing power and cannot meet the requirements of high real-time applications. When Liu et al. \cite{2016_Liu_SSD} developed the Single-shot Multibox Detector (SSD) for the first time, they used convolutional feature layers of different scales to build a feature pyramid in the same neural network and used convolutional feature layers of different scales to detect objects of different scales. SSD utilizes the existing convolutional feature layers in the neural network model to cope with the multi-scale problem of the object without adding additional time and computational cost. Although shallow convolutional features have finer location feature information and can be used to detect small objects, the semantic features of shallow convolutional feature layers are weak, and the effect is not good in predicting small-scale objects. To solve this problem, Sung-Yi Lin et al. \cite{2017_Lin_Feature_Pyramid_Networks} proposed a Feature Pyramid Network (FPN). FPN builds a top-down structure on top of an originally bottom-up network. The model can gradually extract the image's high-level semantic features using the bottom-up structure. Using the top-down structure, the model can fully fuse the high-level semantic features with the shallow position information so that the shallow convolution features also have strong semantic features. Since then, another researcher \cite{2018_Liu_Path_Aggregation_Network} proposed a Path-enhanced Network (PAN). Based on FPN, PAN adds a bottom-up structure, which shortens the transmission path of the underlying accurate positioning information, avoids the loss in the information transmission process, and improves the performance of multi-scale object detection.

The above series of methods have multiple output structures, that is, different scale output structures for different scale ranges of objects. The flying bird object in the surveillance video has a special multi-scale property; in most cases, the scale is small, and in a few cases, the scale is large. Therefore, using the multi-scale output structure to detect the flying bird object, the output structure utilization rate of predicting the large-scale object is low, which will cause unnecessary waste of computing time.

In this paper, we use a large feature layer fusing various resolutions and depth features to detect flying bird objects in surveillance videos.

\subsection{Related Work on Label Assignment}\label{label_assignment}

In the early years, label assignment strategies used fixed and predefined rules. The object detection method based on the anchor box usually uses the Intersection Over Union (IOU) ratio between the anchor box and GT box to realize the assignment of labels. For example, the RPN network \cite{2017_Ren_Faster_RCNN} in Faster R-CNN uses 0.7 and 0.3 as IOU thresholds for positive and negative samples, YOLOV4 \cite{2020_Bochkovskiy_YOLOv4} and YOLOV5 \cite{yolov5_2021} stipulate that the anchor box with the largest IOU value with the GT box is a positive sample. In the object detection method based on anchor-free, some anchor points in the GT box (the feature points of the output feature map) are usually defined as positive samples, and the anchor points outside the GT box are defined as negative samples. For example, FCOS \cite{2019_Tian_FCOS} and Foveabox \cite{2020_Kong_FoveaBox} take the anchor points in the central region of the GT box as positive samples and achieve a relatively ideal detection performance.

The label assignment strategy based on fixed rules is simple and effective, but it does not consider some characteristics of the object (such as shape, size, etc.), so this kind of label assignment strategy is often not optimal. In recent years, researchers have also explored dynamic label assignment strategies. Hengduo Li et al. \cite{2020_Li_Learning_From_Noisy_Anchors} proposed a cleanliness score to indicate the degree of each anchor as a positive sample. Kang Kim et al. \cite{2020_Kim_PAA} proposed a probabilistic anchor assignment strategy (PAA), which uses a Gaussian mixture model to fit the joint distribution of positive and negative samples. Benjin Zhu et al. \cite{2020_Benjin_AutoAssign} explored a fully data-driven way to implement the label assignment method AutoAssign. Zheng Ge et al. \cite{2021_Ge_OTA} first transformed the label assignment problem into an optimization theory problem, regarded the label assignment problem as an optimal transmission problem, and proposed an Optimal Transmission Allocation (OTA) label assignment algorithm. Since then, the team has simplified the OTA algorithm and proposed SimOTA \cite{2021_Zheng_YOLOX}. SimOTA \cite{2021_Zheng_YOLOX} used simple rules instead of the Sinkhorn-Knopp iterative optimization algorithm in OTA, which improved the training speed.

In this paper, we consider using the SimOTA \cite{2021_Zheng_YOLOX} algorithm to allocate labels dynamically. SimOTA \cite{2021_Zheng_YOLOX} determines the number of positive samples by calculating the cumulative Intersection Over Union (IOU) ratio of the GT(Ground Truth) box and each prediction box and uses the cost matrix composed of category loss and IOU loss of the GT box and prediction box to determine the attribution of positive and negative samples. In this paper, we only detect flying bird objects, and there is no problem with multi-categories. The loss function has no category loss, and the SimOTA algorithm cannot be directly used in label assignment. Therefore, based on SimOTA \cite{2021_Zheng_YOLOX}, we further simplify it and propose SimOTA for One-Category object detection (SimOTA-OC) label allocation strategy, which only uses IOU to realize the dynamic allocation of bird object labels.

\section{The Proposed FBOD-SV}\label{The Proposed Methord}

\begin{figure*}[!ht]
\centering
\subfloat{\includegraphics[width=6in]{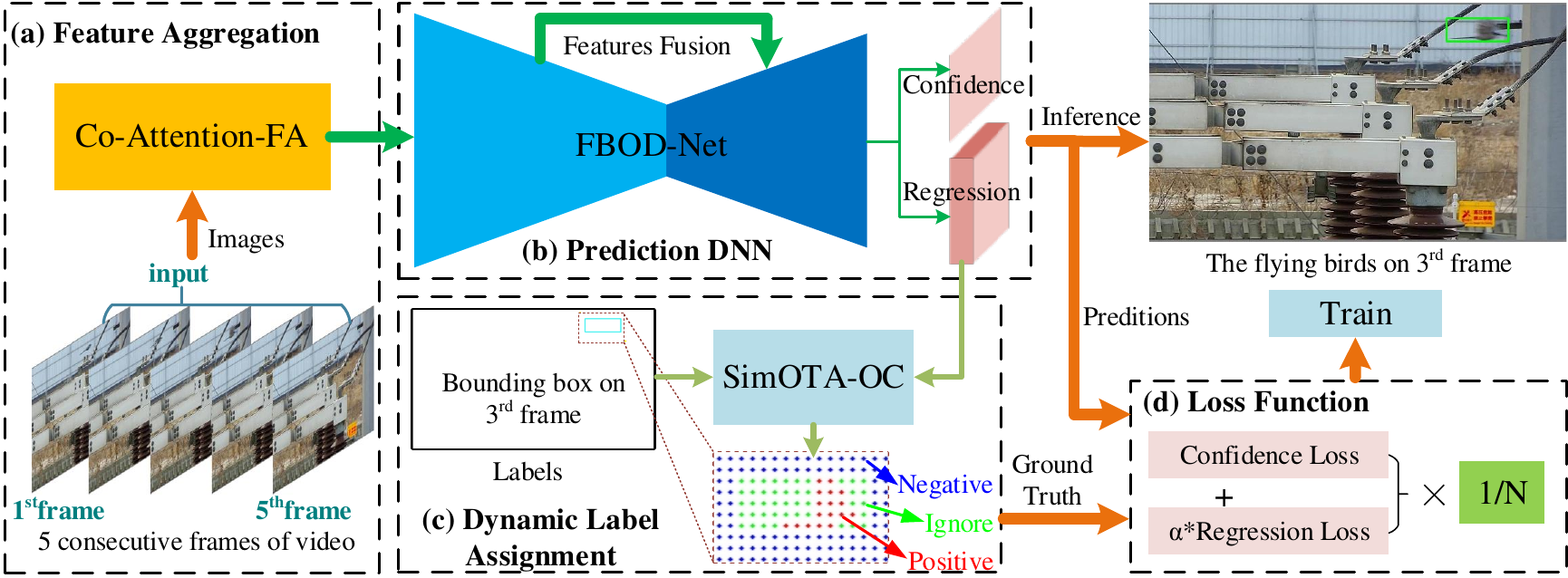}}
\caption{Overview of the proposed FBOD-SV. (a) The Co-Attention-FA unit. (b) The FBOD-Net model. (c) The SimerOTA-OC dynamic label assignment unit. (d) Model training unit (loss function).}
\label{framework_fig}
\end{figure*}

Fig. \ref{framework_fig} shows the summary diagram of the proposed Flying Bird Object Detection method for Surveillance Video (FBOD-SV), which mainly includes four parts, namely, the Co-Attention-FA unit, the FBOD-Net model, the SimOTA-OC label assignment unit, and the model training unit. Among them, the Co-Attention-FA unit aggregates the features of the flying bird object on $n$ consecutive frames of images ($n$=5 as an example in Fig. \ref{framework_fig}). The FBOD-Net model utilizes aggregated features of the flying bird objects to predict the location information of the flying birds in the middle frame of consecutive $n$ frames (the objects in the middle frame of $n$ consecutive frames have symmetric context information). The SimOTA-OC label assignment unit dynamically allocated positive samples according to the model prediction results during model training. The multi-task loss function is used in the model training unit to train the flying bird object detection model. Next, the Co-Attention-FA unit (\ref{method_a}), the FBOD-Net (\ref{method_b}), the SimOTA-OC dynamic label assignment method (\ref{method_c}), and the loss function (\ref{method_d}) will be introduced, respectively.

\subsection{The Co-Attention-FA}\label{method_a}

Most of the flying bird objects are not obvious in a single frame image, so we aggregate the characteristics of the flying bird objects on consecutive frames of images to enhance the flying bird objects' characteristics and improve the detection rate of the flying bird objects. In particular, inspired by the spatial attention mechanism \cite{2015_JaderBerg_STN}, we propose a new method to aggregate the information of consecutive multiple frames, the The Co-Attention-FA method, which uses the information of other frames to guide the current frame to pay attention to and extract the features of the flying bird object in the case that the characteristics of the bird object cannot be extracted in a single frame.

\begin{figure*}[!ht]
\centering
\includegraphics[width=6in]{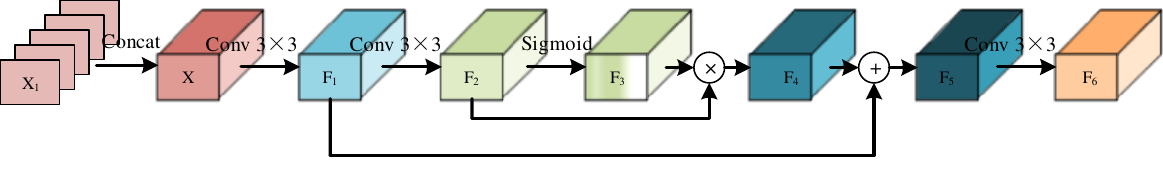}
\caption{Diagram of the Co-Attention-FA module.}
\label{correlevant_attention_fig}
\end{figure*}

The diagram of the Co-Attention-FA module is shown in Fig. \ref{correlevant_attention_fig}. Firstly, the consecutive frames of images $ \left\{ \text{x}_1, \text{x}_2,..., \text{x}_n\right\}$ are concated into a multi-channel feature matrix $\text{X}$, prepared for the information correlation between frames. Then, the information of consecutive frames is correlated and fused by $3\times3$ convolution to obtain the feature matrix $\text{F}_1$. At this time, $\text{F}_1$ already has the correlation information between frames.

Next is a residual structure. Inside the residual structure, $\text{F}_1$ is first carried out $3\times3$ convolution operation to get $\text{F}_2$, and the correlation information between frames is used to make $\text{F}_2$ focus on the bird object. Then $\text{F}_2$ is operated by the Sigmoid function to obtain $\text{F}_3$ to make the focusing information explicit. $\text{F}_4$ is obtained by dot multiplication of $\text{F}_3$ and $\text{F}_2$ to realize spatial attention.

On the outside of the residual structure, $\text{F}_4$ is superimposed on $\text{F}_1$. The purpose of such processing is to highlight the characteristics of the flying bird object while retaining the necessary background information. Finally, the aggregated feature $\text{F}_6$ is output by a $3\times3$ convolution operation.

The flying bird may have different positions and shapes on consecutive video frames. However, the spatial position of the flying bird across $n$ consecutive frames of images is continuous, and its action is coherent [as shown in Fig. \ref{Flying_bird_in_surveillance_videos}\subref{obvious_bird_fig}]. Compared to a single frame, the flying bird object exhibits richer features on consecutive frames of images. Therefore, leveraging the characteristics of Co-Attention-FA, we can train the model using backpropagation and gradient descent algorithms to correctly aggregate the features of the flying bird objects across the consecutively $n$ frames of the image.

\subsection{The FBOD-Net}\label{method_b}

The flying birds in surveillance videos have special multi-scale properties (i.e., most are small objects), so we adopt a large feature map to predict the flying birds in surveillance videos. However, the shallow large feature maps do not have a large receptive field, which is not conducive to detecting large objects. At the same time, the semantic information of the shallow large feature maps is weak, which is not conducive to object recognition. Therefore, we fuse the information of the deep small feature map of Deep Neural Network (DNN) into the large feature map so that the large feature map has both delicate spatial information and abstract high-level semantic information.

\begin{figure}[!ht]
\centering
\includegraphics[width=3.2in]{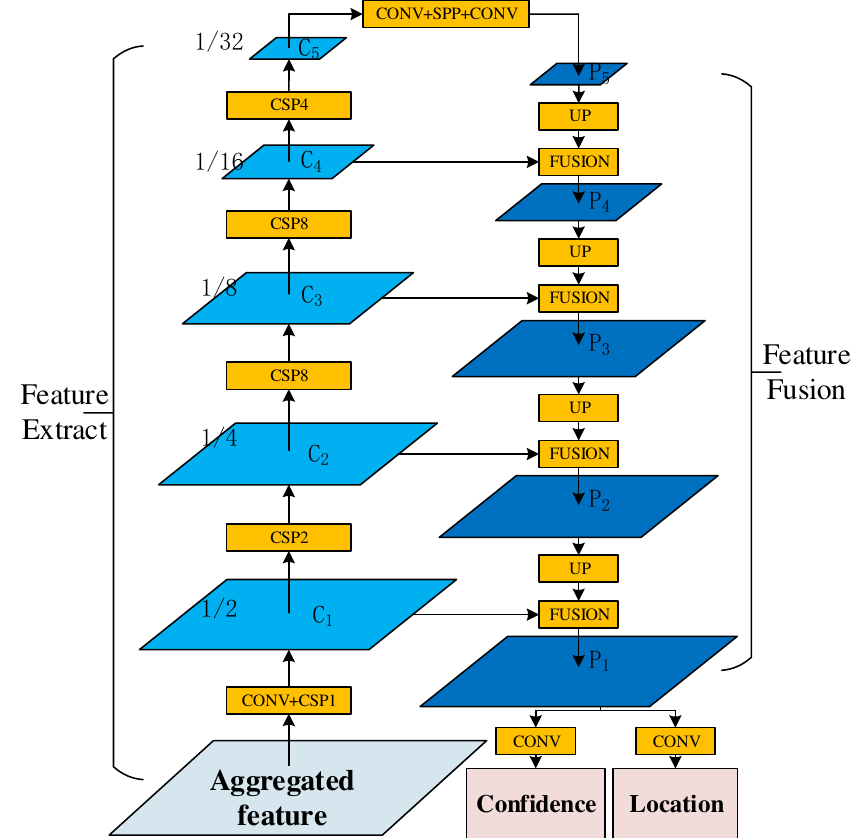}
\caption{The FBOD-Net.}
\label{FBOD_Net_fig}
\end{figure}

\begin{figure}[!ht]
\centering
\includegraphics[width=1.5in]{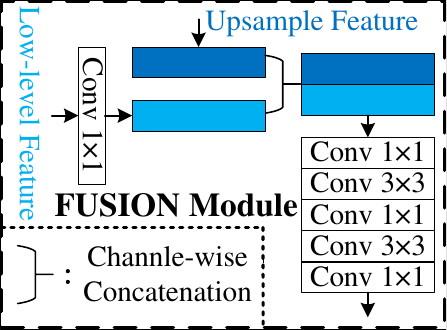}
\caption{Feature fusion module in FBOD-Net.}
\label{Fusion_module_fig}
\end{figure}

Specifically, we adopt a network structure with down-sampling followed by up-sampling to fully fuse the shallow and deep feature map information, as shown in Fig. \ref{FBOD_Net_fig}. Firstly, CSPDarkNet53 \cite{2020_Wang_CSPNet} was used to extract the down-sampling style features of the feature aggregated image, and a series of feature maps ($\text{C}_1$, $\text{C}_2$, $\text{C}_3$,$\text{C}_4$, $\text{C}_5$ in Fig. \ref{FBOD_Net_fig}) were obtained. Then, the CONVolution (CONV) + Spatial Pyramid Pooling (SPP)+ CONV operation is performed on $\text{C}_5$ to obtain the feature map $\text{P}_5$. $\text{P}_5$ is upsampled (the bilinear interpolation method is used in this paper) and then fused with $\text{C}_4$ (a schematic of the fusion operation is shown in Fig. \ref{Fusion_module_fig}) to obtain the feature map $\text{P}_4$. The feature layers $\text{P}_4$, $\text{P}_3$, and $\text{P}_2$ perform the same operation as $\text{P}_5$ (upsampled and then fused with the shallow feature map of the same scale) and finally obtain the feature map $\text{P}_1$. Finally, the feature map $\text{P}_1$ undergoes two different convolution operations simultaneously to obtain the feature map of prediction confidence and the feature map of regression flying bird object position.

\subsection{SimOTA-OC Dynamic Label Assignment Method}\label{method_c}

Since the flying bird objects to be detected in the surveillance video are mostly small objects, if the preset anchor box is used to detect the flying bird objects, most of the larger anchor boxes cannot be used and are redundant. The object detection method based on anchor-free does not need a preset box and directly uses the feature points (anchor points) of the feature map to predict the category and location information of the object, which is simple and clear. Therefore, we adopt an anchor-free approach to detect flying bird objects.

Some anchor-free methods, such as Foveabox \cite{2020_Kong_FoveaBox} and CenterNet \cite{2019_Duan_CenterNet}, focus on the center of the bounding box when assigning labels. However, in the surveillance video, the flying bird objects in the bounding box are mostly asymmetric, and the center of the bounding box cannot represent the flying bird objects well (as shown in Fig. \ref{Flying_bird_in_surveillance_videos}\subref{irregularity_scale_bird_fig}). In addition, this kind of label allocation method uses a static allocation strategy. In some cases, the boundary of the birds in the surveillance video is fuzzy, and it is difficult to manually find the boundary between positive and negative samples. Therefore, these methods are unsuitable for label assignment strategies for training bird object detection models. Some other methods based on anchor-free adopt dynamic label assignment (such as OTA \cite{2021_Ge_OTA}, SimOTA \cite{2021_Zheng_YOLOX}) to determine the assignment problem of labels, which can well solve the problem of attribute (positive and negative samples) and attribution (which truth object the sample belongs to) of samples. OTA \cite{2021_Ge_OTA} is complex and will prolong the training time of the model. However, SimOTA \cite{2021_Zheng_YOLOX} is a simplified version of OTA \cite{2021_Zheng_YOLOX}, which can reduce the training time by about 20\%-25\% while ensuring the accuracy of model detection. So, we refer to the SimOTA \cite{2021_Zheng_YOLOX} method when assigning labels.

SimOTA \cite{2021_Zheng_YOLOX} determines the number of positive samples by calculating the cumulative Intersection Over Union (IOU) ratio of the GT(Ground Truth) box and each prediction box and uses the cost matrix composed of category loss and IOU loss of the GT box and prediction box to determine the attribution of positive and negative samples. In this paper, we only detect flying bird objects, and there is no problem with multi-categories. The loss function has no category loss, and the SimOTA algorithm cannot be directly used in label assignment. Therefore, based on SimOTA \cite{2021_Zheng_YOLOX}, we further simplify it and propose SimOTA for One-Category object detection (SimOTA-OC) label allocation strategy, which only uses IOU to realize the dynamic allocation of bird object labels. Specifically, SimOTA-OC has the following steps (We preset anchor points in the GT box as positive samples and anchor points outside the GT box as negative samples. The following steps only operate on the preset positive samples; the anchor points not judged as positive samples are treated as negligible samples).

Firstly, the IOU of each anchor point prediction box and each GT box are calculated to form an IOU matrix (as shown in TABLE \ref{tab:IOU_matrix}).

\begin{table}[!ht]
\caption{IOU matrix of GT boxes and anchor prediction boxes.\label{tab:IOU_matrix}}
\centering
\begin{tabular}{c|c c c c}
\hline
\quad & $\text{A}_{1}$ & $\text{A}_{2}$  & ... & $\text{A}_{n}$ \\
\hline
$\text{GT}_1$  & $\text{IOU}_{11}$ & $\text{IOU}_{12}$ & ... & $\text{IOU}_{1n}$ \\
$\text{GT}_2$  & $\text{IOU}_{21}$ & $\text{IOU}_{22}$ & ... & $\text{IOU}_{2n}$ \\
...  & ... & ... & ... & ... \\
$\text{GT}_m$  & $\text{IOU}_{m1}$ & $\text{IOU}_{m2}$ & ... & $\text{IOU}_{mn}$ \\
\hline
\end{tabular}
\end{table}

Then, for a $\text{GT}$ box (with $\text{GT}_{m}$, for example), the accumulated IOU value of the row of $\text{GT}_{m}$ in the IOU matrix is calculated and rounded upward as the number of positive sample anchor points corresponding to $\text{GT}_{m}$,
\begin{align}
{\text{P}}_{m} = \text{ceil}\left ( \sum_{i=1}^{n}\text{IOU}_{mi} \right ),
\end{align}
where ${\text{P}}_{m}$ means the number of positive sample anchor points of $\text{GT}_{m}$, and $\text{ceil}\left ( \cdot \right )$ denotes the upper rounding function.

Finally, the IOU values of $\text{GT}_{m}$ and all anchor point prediction boxes are sorted from large to small, and the top ${\text{P}}_{m}$ anchor points are taken as the positive samples of GT. If two or more GTs own an anchor point simultaneously, it will be assigned to the GT with a large IOU value. If the IOU values are all equal, the anchor point will be treated as an ignored sample.

\subsection{Loss Function}\label{method_d}
This paper uses a multi-task loss function to train the bird object detection model. Since only flying bird objects are detected, and there are no objects of other classes, no classification loss is applied. The multi-task loss in this paper includes confidence loss and location regression loss. If the anchor point is a positive sample, the loss is the weighted sum of the confidence loss and the location regression loss; if the anchor point is a negative sample, then its loss only contains confidence loss; if the anchor point is an ignored sample, then its loss value is 0 (does not participate in backpropagation),
\begin{align}
\text{L}\left (\text{A}_i\right)=
\begin{cases}
     {\text{L}}_{\text{Conf}}\left (\text{A}_i\right) + {\alpha}{\text{L}}_{\text{Reg}}\left (\text{A}_i\right), &{\text {if }} \text{A}_i {\text { is}} {\text { Positive}},\\
    {\text{L}}_{\text{conf}}\left (\text{A}_i\right), &{\text {if }} \text{A}_i {\text { is}} {\text { Negative}},\\
    0, &{\text { Otherwise}},
\end{cases}
\end{align}
where ${\text{L}}_{\text{Conf}}\left (\cdot\right)$ and ${\text{L}}_{\text{Reg}}\left (\cdot\right)$ represent confidence loss location and regression loss, respectively, and $\alpha$ means the balance parameter of the two kinds of loss. The L2 loss is used for the confidence loss, and the CIOU \cite{2021_zheng_CIOU} loss is used for the location regression loss. The loss of a single anchor point can also be expressed as follows,
\begin{align}
\text{L}\left (\text{A}_i\right)={\text{L}}_{\text{Conf}}\left (\text{A}_i\right) + {\alpha}{\text{L}}_{\text{Reg}}\left (\text{A}_i\right),
\end{align}
\begin{align}
{\text{L}}_{\text{Conf}}\left (\text{A}_i\right)=
\begin{cases}
     \left\|{\text{Conf}_{\text{pred}}}-1\right\|_2, &{\text {if }} \text{A}_i {\text { is}} {\text { Positive}},\\
    \left\|{\text{Conf}_{\text{pred}}}\right\|_2, &{\text {if }} \text{A}_i {\text { is}} {\text { Negative}},\\
    0, &{\text { Otherwise}},
\end{cases}
\end{align}
\begin{align}
&{\text{L}}_{\text{Reg}}\left (\text{A}_i\right)=\notag\\
&\begin{cases}
     {\text{LCIOU}}\left({\text{Box}_\text{pred}}\left (\text{A}_i\right),{\text{Box}_{\text{GT}}}\left (\text{A}_i\right)\right), &{\text {if }} \text{A}_i {\text { is}} {\text { Positive}},\\
    0, &{\text { Otherwise}},
\end{cases}
\end{align}
Where ${\text{Conf}_\text{pred}}$ represents the confidence prediction value of anchor point $\text{A}_i$, $\text{LCIOU}\left (\cdot\right)$ means CIOU loss, ${\text{Box}_\text{pred}}$ is the bounding box predicted by anchor point $\text{A}_i$ for the flying bird object to which the anchor point belongs, and ${\text{Box}_\text{GT}}$ represents the Ground Truth bounding box for the flying bird object to which the anchor point $\text{A}_i$ belongs. The total loss is equal to the sum of all anchor point losses,
\begin{align}
\text{Total Loss} &=\frac{1}{\text{N}} \sum{\text{L}\left (\text{A}_i\right)}\notag\\
&=\frac{1}{\text{N}}\left({\sum{\text{L}}_{\text{Conf}}\left (\text{A}_i\right)} + {\alpha}\sum{{\text{L}}_{\text{Reg}}\left (\text{A}_i\right)}\right)\notag\\
&=\frac{1}{\text{N}}\left({\text{L}}_{\text{Conf}}+{\alpha}{\text{L}}_{\text{Reg}} \right),
\end{align}
where $\text{N}$ is the normalization parameter. $\text{N}$ is the number of positive anchor point samples when the image contains flying bird objects, and a fixed positive number when the image does not contain flying bird objects. ${\text{L}}_{\text{Conf}}$ is the confidence loss for all anchor points, and ${\text{L}}_{\text{Reg}}$ is the regression loss for all anchor points.

\section{Experiment}\label{Experiment}

In this part, quantitative and qualitative experiments will be conducted to demonstrate the effectiveness of the proposed FBOD-SV. Next, we will describe the dataset (\ref{datasets}), evaluation method (\ref{evaluation_metrics}), experimental platform (\ref{experimental_platforms}), implementation details (\ref{implementation_details}), parameter analysis experiment (\ref{parameter_analysis}), ablation comparative analysis experiments (\ref{ablation_analysis}), and other methods comparative analysis experiments (\ref{comparative_analysis}).

\subsection{Datasets}\label{datasets}

\begin{figure}[!ht]
\centering
\includegraphics[width=3.2in]{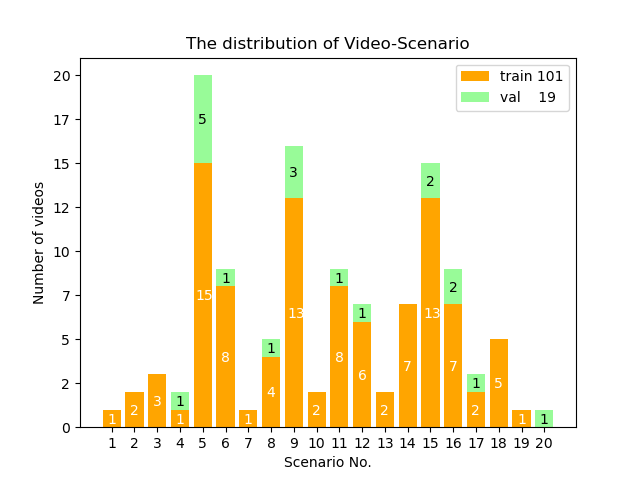}
\caption{The distribution of Video-Scenario.}
\label{video_scenario_distribution}
\end{figure}

\begin{figure}[!ht]
\centering
\includegraphics[width=3.2in]{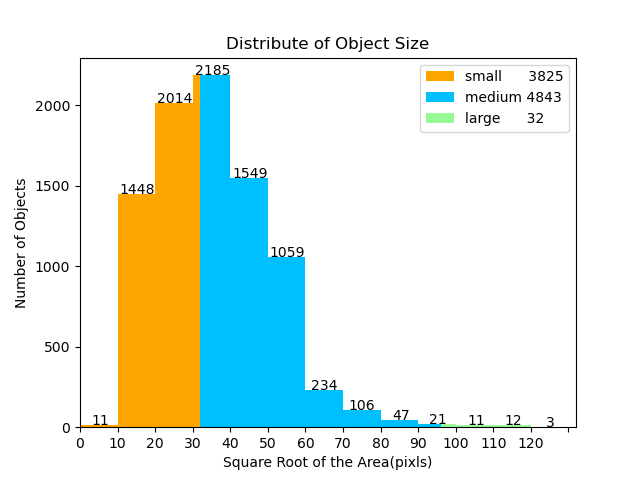}
\caption{Size distribution of the flying birds in the datasets.}
\label{size_curve_fig}
\end{figure}

Based on our previous work \cite{2024_sun_Flying_Bird}, we collected five videos containing flying birds (the size of the video images is 1280$\times$720) in an unattended traction substation again. One of the videos is randomly selected to be expanded to the test set, and the remaining four videos are expanded to the training set (at this time, there are 101 videos in the training set and 19 videos in the test set, for a total of 120 videos). We end up with 28353 images, 7736 of which contain flying birds, for a total of 8700 flying bird objects. The training set includes 24898 images, and the test set includes 3455 images. At the same time, we checked and modified all the annotated bounding boxes. Fig. \ref{video_scenario_distribution} and Fig. \ref{size_curve_fig} are the Video-Scenario distribution and the size distribution of the flying birds after the dataset is expanded based on \cite{2024_sun_Flying_Bird}.

\subsection{Evaluation Metrics}\label{evaluation_metrics}

Same as \cite{2024_sun_Flying_Bird}, average precision (AP), the evaluation metric of Pascal VOC 2007 \cite{2010_Pascal_VOC}, is used to evaluate the detection results of the model. Specifically, $\text{AP}_{50}$ (The subscript 50 indicates that a test result is considered a true positive when the IOU between the test result and the true value is greater than or equal to 50\%. That is, the IOU threshold is set to 50\%), $\text{AP}_{75}$ (subscript 75 has a similar meaning to subscript 50), and AP (average accuracy exceeds multiple threshold averages, and the IOU threshold is set from 50\% to 95\% with 5\% interval).

\subsection{Experimental Platforms}\label{experimental_platforms}

All the experiments are implemented on a desktop computer with an Intel Core i7-9700 CPU, 32 GB of memory, and a single NVIDIA GeForce RTX 3090 with 24 GB GPU memory.

\subsection{Implementation Details}\label{implementation_details}

\begin{figure}[htb]
\centering
\subfloat[]
{\includegraphics[width=3.2in]{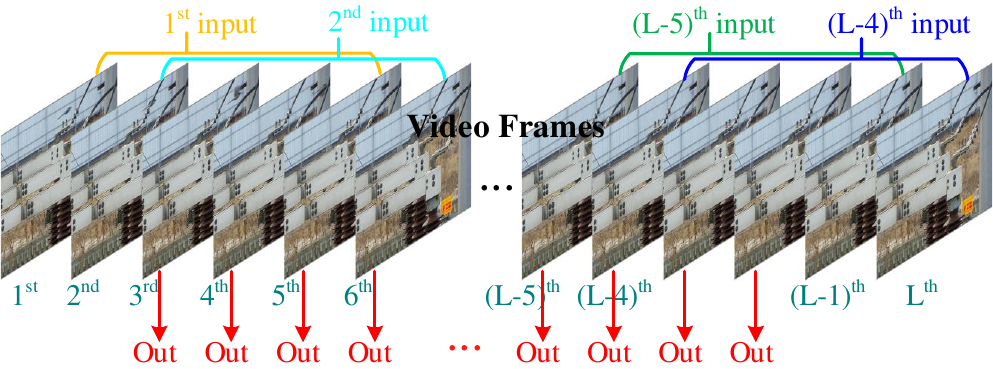}
\label{Original}}
\\
\subfloat[]
{\includegraphics[width=3.2in]{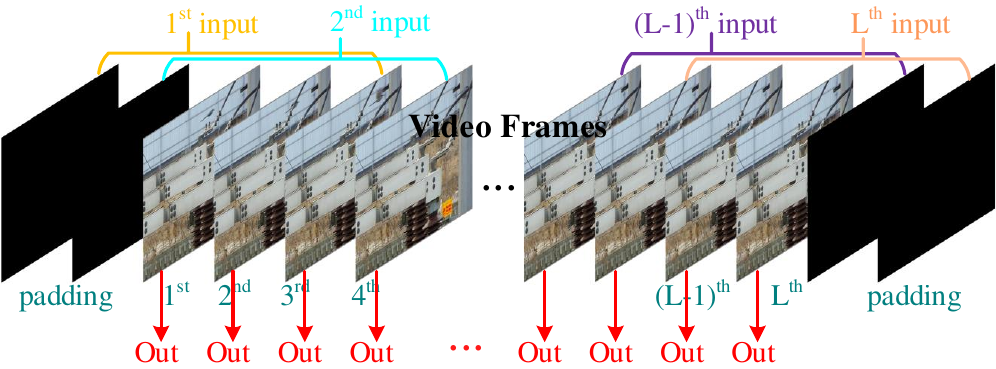}
\label{After_padding}}
\caption{Schematic diagram of input and output for detecting flying birds in a video using FBOD-SV. (a) No padding operation is applied to the video image sequence. (b) The padding operation is performed on the video image sequence.}
\label{detection_in_video}
\end{figure}

Input $n$ consecutive 3-channel RGB images of size 672$\times$384 into the Co-Attention-FA, and output a feature aggregated image with ($n$$\times$3+1) channels and a size of 672$\times$384. The input of the FBOD-Net is the output of the Co-Attention-FA, and the output is a 336$\times$192$\times$1 confidence prediction feature map and a 336$\times$192$\times$4 position regression feature map. The output predicts the position of the object on the intermediate frame. Since only the object of the middle frame of consecutive $n$ frames is predicted when the input is a video, the first $(n-1)/2$ frames and the last $(n-1)/2$ frames have no detection result, as shown in Fig. \ref{detection_in_video}\subref{Original} ($n$=5 as an example). To solve this problem, this paper introduces video image sequence padding according to the principle of convolutional padding operation. A schematic diagram of the padding operation of the video image sequence is shown in Fig. \ref{detection_in_video}\subref{After_padding}, where an all-black image (each pixel of the image has a value of 0) is extended by $(n-1)/2$ frames at the beginning and end of the video image sequence. After the padding operation of the video image sequence, each frame of the original video sequence will have the prediction result.

In this paper, all experiments are implemented under the Pytorch framework. All network models are trained on an NVIDIA GeForce RTX 3090 with 24 GB of GPU memory. For the batch size setting, the training model designed in this paper is set to 8, and other comparison experimental models are set according to their model size. All experimental models were trained from scratch without pre-trained models. The initial learning rate is set to 0.001, and the ratio decreases during training (per iteration, multiplied by 0.95). In addition, when training the model designed in this paper, based on \cite{2024_sun_Flying_Bird}, data augmentation adds enhancement operations such as random Cropping, random Center Flipping, and random HSV (All data augmentation performs the same operation on $n$ consecutive frames, so Mosaic and Mixup data augments are inappropriate). The model is trained for a total of 100 iterations.

\subsection{Parameter Analysis Experiment}\label{parameter_analysis}

In this subsection, we conduct analysis experiments on the key parameters that affect the algorithm's performance. Specifically, we analyze the effect of different numbers of consecutive input frames (the parameter $n$ mentioned above) on the algorithm's performance. Three different numbers of consecutive input frames are set to test the influence on the algorithm's performance. In particular, the effect of 3, 5, and 7 consecutive input frames on the algorithm's performance is tested. In theory, the more consecutive input frames there are, the more information it can obtain. However, for the proposed method, as the number of consecutive input frames increases, the difficulty of information fusion will also increase, and the algorithm's detection accuracy will not necessarily grow. In addition, the model inference time increases as the number of consecutive input frames increases.

The experimental results of the influence of different consecutive input frames on the algorithm's performance are shown in TABLE \ref{tab:Effect_Diff_Input_Frames}. The results show that when the number of consecutive input frames is 3, the algorithm's accuracy is the lowest, but the running speed of the model is the fastest. The algorithm's accuracy is the highest when the number of consecutive input frames is 5. When the number of consecutive input frames is 7, the running speed of the model is slow, and the algorithm's accuracy is not the highest. Considering comprehensively, the number of consecutive input frames of the model is set to 5 in the process of subsequent experiments.

\begin{table}[!ht]
\caption{Effect of different numbers of consecutive input frames on the performance of the algorithm.}\label{tab:Effect_Diff_Input_Frames}
\centering
\begin{tabular}{c|c c c c}
\hline
Frames & $\text{AP}_{50}$ & $\text{AP}_{75}$  & AP & \makecell[c]{Inference\\Time (s)} \\
\hline \hline
3  & 0.731 & 0.298 & 0.305 & 0.00697\\
5 & 0.762 & 0.371 & 0.395 & 0.00704\\
7 & 0.760 & 0.369 & 0.393 & 0.00712\\
\hline
\end{tabular}
\end{table}

\subsection{Ablation Comparative Analysis Experiment}\label{ablation_analysis}

In this subsection, three ablation contrast experiments will be set up according to the solution method of the challenges faced by detecting the flying bird object in the surveillance video to prove the effectiveness of the proposed solution. In particular, there are comparison experiments between single-frame image input and continuous multiple-frame image input \ref{ablation_analysis_1}, between multi-scale model structure and single-scale model structure \ref{ablation_analysis_2}, and between different label assignment strategies \ref{ablation_analysis_3}.

\subsubsection{Comparison Experiment Between Single Frame Image Input and Continuous Multiple Frame Image input}\label{ablation_analysis_1}

We set up a comparison experiment between single-frame image input and continuous multiple-frame image input to show that aggregated features are necessary for flying bird objects in surveillance videos. Specifically, a single image input means that a single image is used as the input to the model for both training and testing. Continuous multi-frame image input means that feature aggregation is performed on multiple frames of images at the input of the detection network, and then the aggregated features are input into the detection network for feature extraction and feature fusion. Among them, two types of aggregation methods are compared for continuous multi-frame image input: the aggregation method based on ConvLSTM \cite{2024_sun_Flying_Bird} and the aggregation method based on Co-Attention-FA. Except for the input method, the model architecture, loss function, label assignment method, and training method are all the same.

\begin{figure*}[!htp]
    \centering
    \subfloat{
        \begin{minipage}[t]{0.225\linewidth}
        \centering
        \includegraphics[width=1\linewidth]{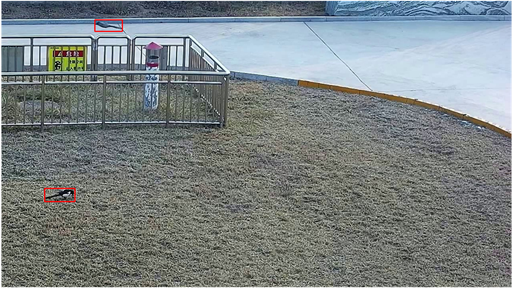}
        \end{minipage}
        }
    \subfloat{
        \begin{minipage}[t]{0.225\linewidth}
        \centering
        \includegraphics[width=1\linewidth]{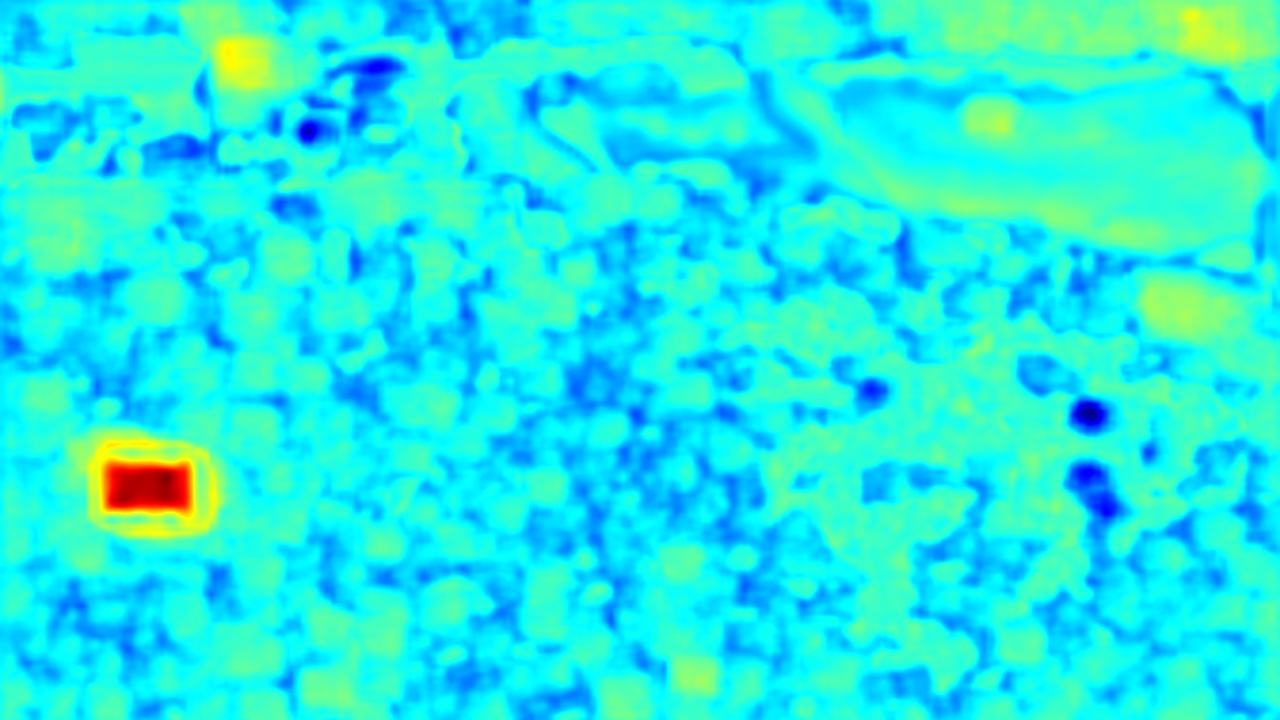}
        \end{minipage}
        }
    \subfloat{
        \begin{minipage}[t]{0.225\linewidth}
        \centering
        \includegraphics[width=1\linewidth]{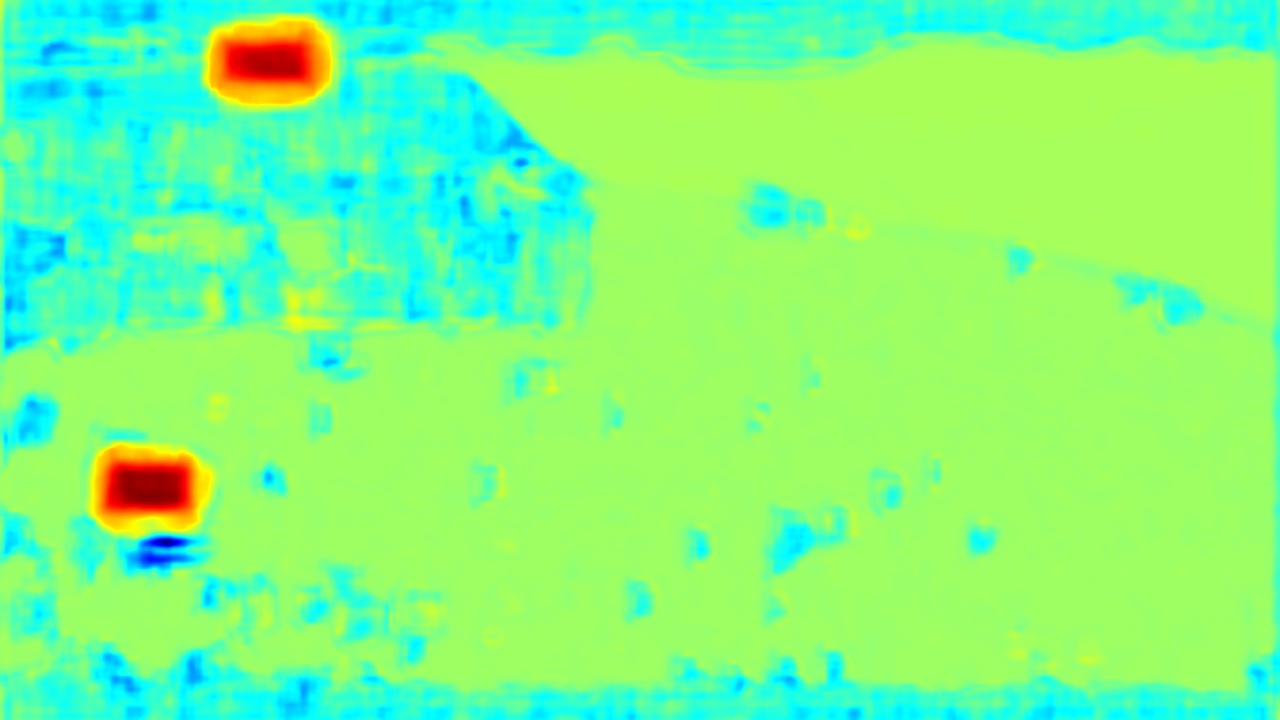}
        \end{minipage}
        }
    \subfloat{
        \begin{minipage}[t]{0.225\linewidth}
        \centering
        \includegraphics[width=1\linewidth]{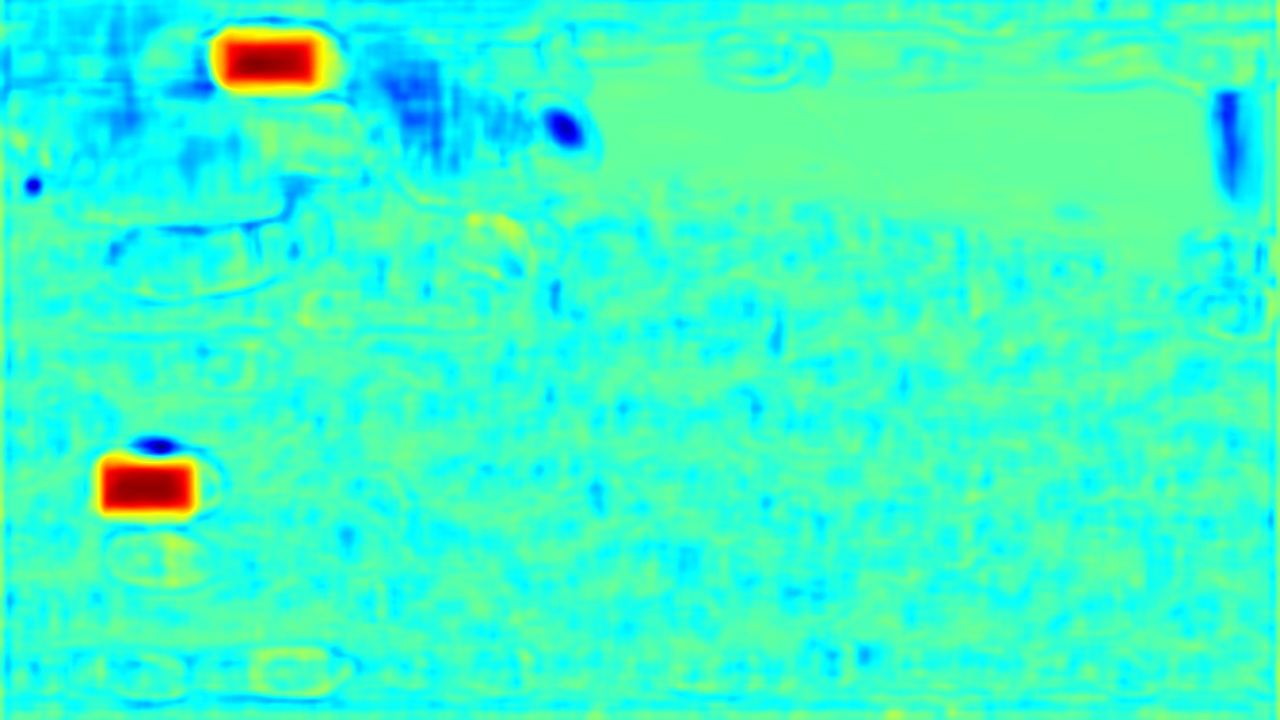}
        \end{minipage}
        }
    
    \vspace{-1mm}
	\setcounter{subfigure}{0}
    
    \subfloat[Raw Image]{
        \begin{minipage}[t]{0.225\linewidth}
        \centering
        \includegraphics[width=1\linewidth]{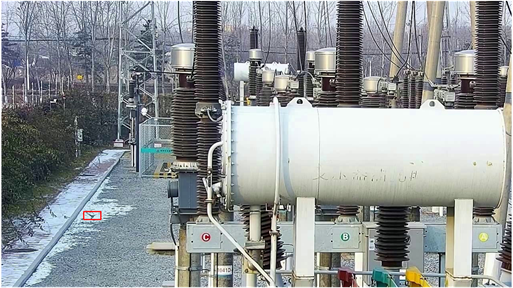}
        \label{Raw_Image}
        \end{minipage}
        }
    \subfloat[w/o Aggregation]{
        \begin{minipage}[t]{0.225\linewidth}
        \centering
        \includegraphics[width=1\linewidth]{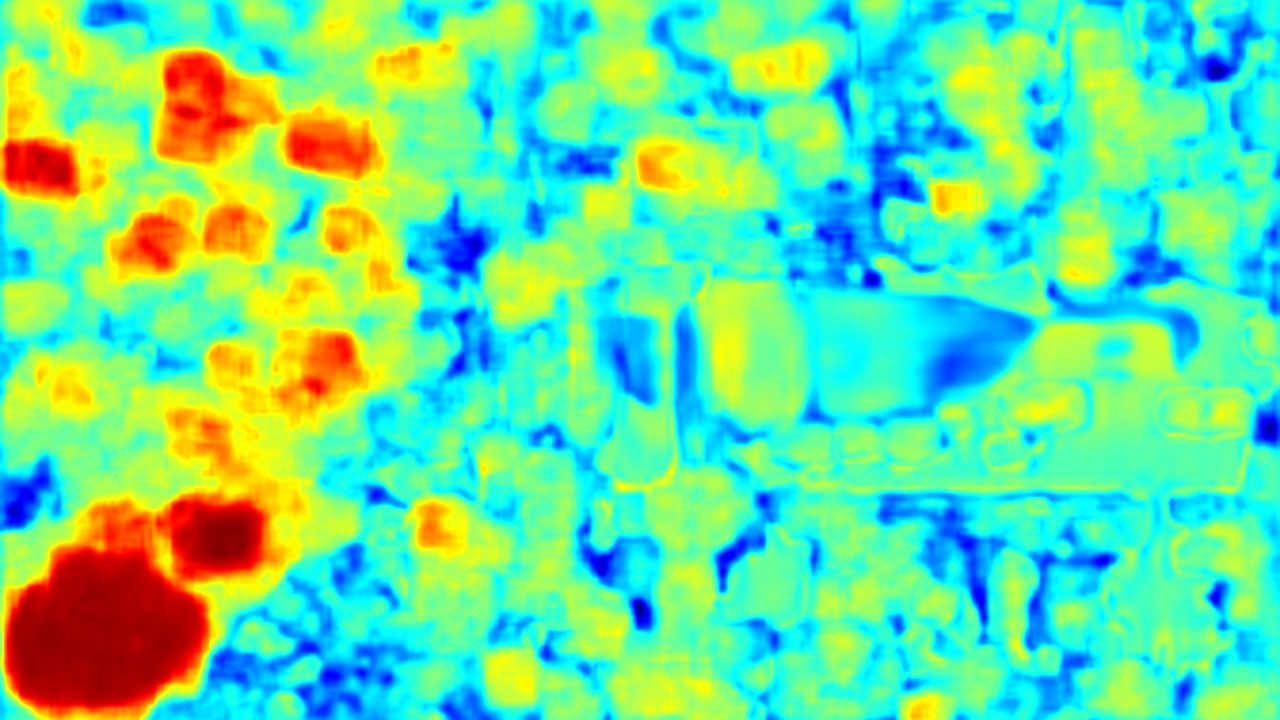}
        \label{w/o_Aggregation}
        \end{minipage}
        }
    \subfloat[ConvLSTM]{
        \begin{minipage}[t]{0.225\linewidth}
        \centering
        \includegraphics[width=1\linewidth]{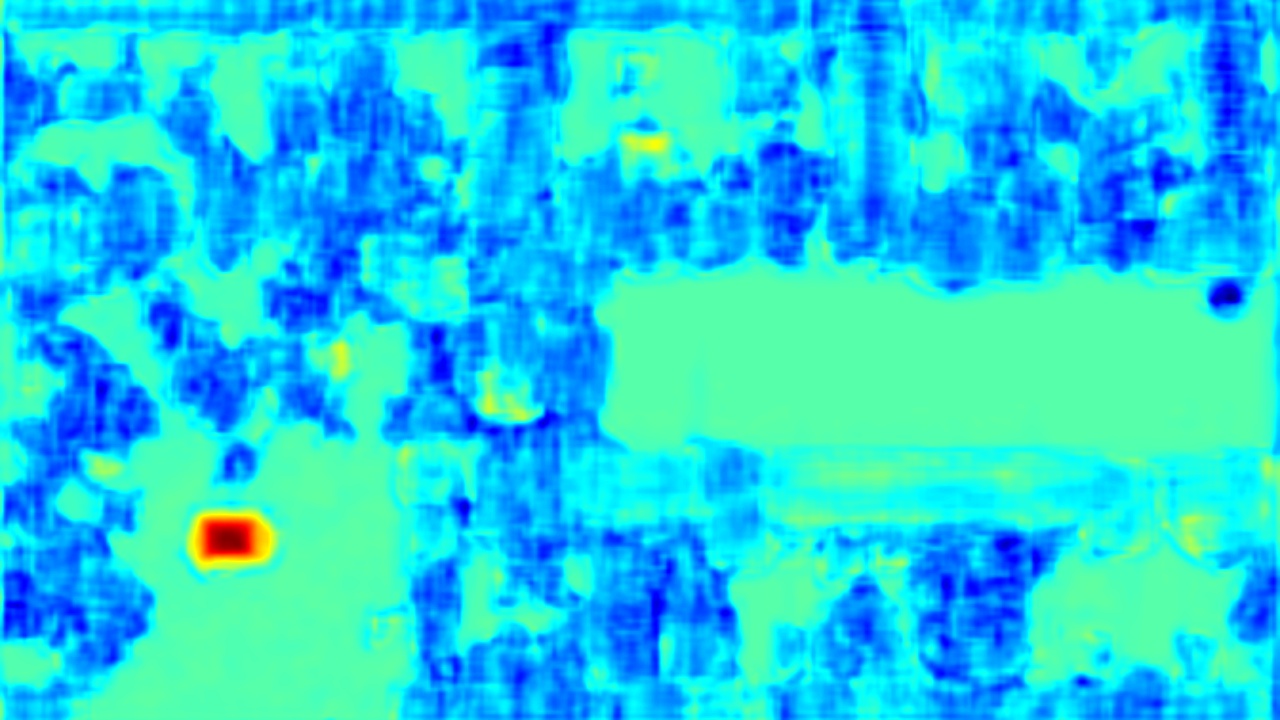}
        \label{ConvLSTM}
        \end{minipage}
        }
    \subfloat[Co-Attention-FA]{
        \begin{minipage}[t]{0.225\linewidth}
        \centering
        \includegraphics[width=1\linewidth]{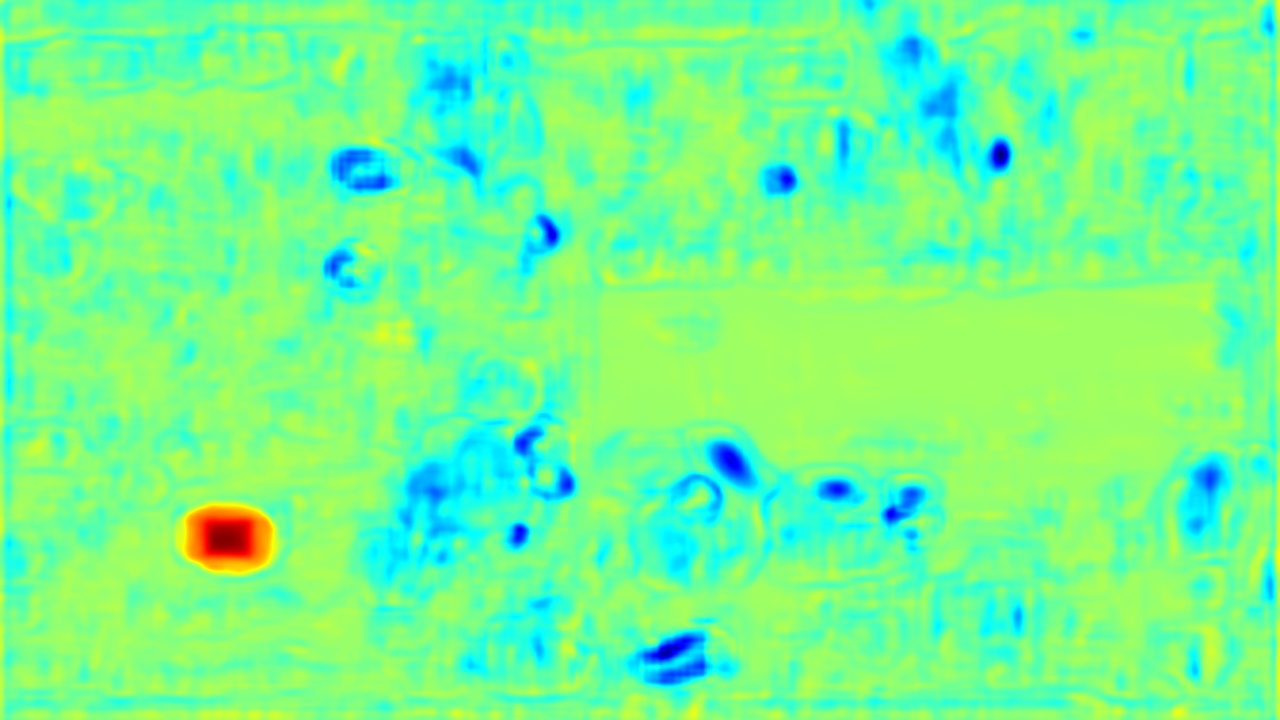}
        \label{Co-Attention-FA}
        \end{minipage}
        }
    \caption{The Heatmaps of image features extracted by models using different feature aggregation methods. (b) Heatmap of image features extracted by the model without feature aggregation method (single frame image input). (c) Heatmap of image features extracted by the model based on ConvLSTM aggregation method (5 consecutive frames of image input). (d) Heatmap of image features extracted by the model based on the Co-Attention-FA aggregation method (5 consecutive frames of image input).}
    \label{Feature_Heatmap}
\end{figure*}

In most cases, the characteristics of the flying bird in a single frame image are not obvious, and the features of the flying bird object cannot be extracted, so it is difficult to detect the flying bird object in the surveillance video. By aggregating the features of consecutive multi-frame images, the features of the flying bird object can be enriched so the flying bird object can be detected. Fig. \ref{Feature_Heatmap} shows the heatmaps of image features extracted by the models with different feature aggregation methods. It can be seen from the figure that after the feature aggregation of consecutive multiple frames of images, the features of the flying bird object, which are not obvious in the single-frame image, can be extracted.

The experimental results of the quantitative comparison are shown in TABLE \ref{tab:Effect_Diff_NUM}. Compared with the single-frame image input, the $\text{AP}_{50}$ of continuous multi-frame image input (aggregation method based on ConvLSTM \cite{2024_sun_Flying_Bird}) is increased by about 1.5 times, reaching 73.3\%, which further supports the above view that continuous multi-frame feature aggregation can enhance the features of bird objects. Compared with the ConvLSTM-based aggregation method \cite{2024_sun_Flying_Bird}, the $\text{AP}_{50}$ of the aggregation method based on Co-Attention-FA is increased by 2.9\%, and the model inference speed by 10.26\%, which shows that the feature aggregation method based on Co-Attention-FA proposed in this paper is more advantageous.

\begin{table}[!ht]
\caption{Effect of different feature aggregation methods on algorithm performance.\label{tab:Effect_Diff_NUM}}
\centering
\begin{threeparttable}
\begin{tabular}{c|c|c c c c}
\hline
Frames  & Aggregation Method & $\text{AP}_{50}$ & $\text{AP}_{75}$  & AP  & \makecell[c]{Inference\\Time (s)} \\
\hline \hline
1\tnote{1}  & w/o  & 0.484 & 0.224 & 0.141 & 0.0069\\
5  & ConvLSTM  & 0.733 & 0.184 & 0.319 & 0.0078\\
5  & Co-Attention-FA  & 0.762 & 0.371 & 0.395 & 0.0070 \\
\hline
\end{tabular}
\begin{tablenotes}
        \footnotesize
        \item[1] To maintain consistency with the conditions of the other two comparison items,  Mosaic and Mixup image enhancements were not applied to the single-frame image input. Therefore, its accuracy will be slightly lower than other advanced image-based object detection methods.
\end{tablenotes}
\end{threeparttable}
\end{table}

\subsubsection{Comparison Experiment Between Multi-scale Model Structure and Single-scale Model Structure}\label{ablation_analysis_2}

\begin{figure}[!ht]
\centering
\includegraphics[width=3.45in]{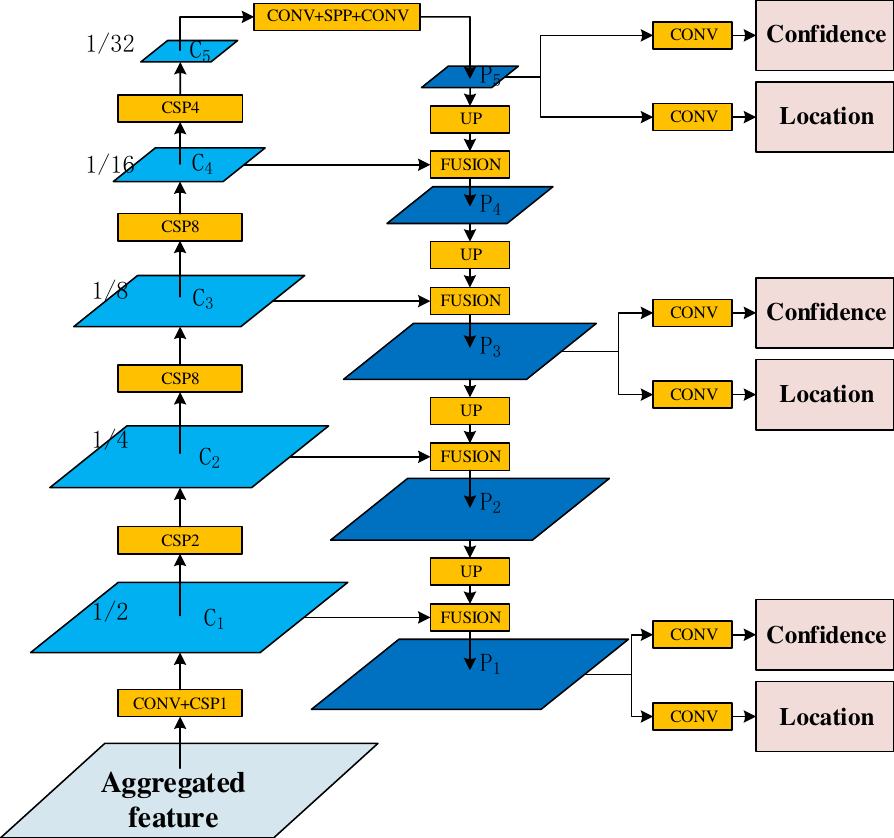}
\caption{The multi-scale output flying bird object detection structure corresponds to the proposed single-scale output bird-object detection model.}
\label{multi-output_net_fig}
\end{figure}

Flying birds in surveillance videos have special multi-scale properties. Specifically, they belong to small objects in most cases. Therefore, it is not necessary to adopt a multi-scale structure to detect flying birds in surveillance videos, and a feature layer with a larger scale that fuses multi-scale information is used for detection in this paper. To verify that the single-scale model structure designed in this paper is superior to the multi-scale model structure for detecting flying birds in surveillance videos, we design the corresponding multi-scale model structure (structure diagram shown in Fig. \ref{multi-output_net_fig}) for comparison experiments with the single-scale model structure. In particular, three scales are designed to detect multi-scale bird objects. We use the K-means clustering method for its object scale assignment, whose input, loss function, label assignment method, and training method are completely consistent with the one in this paper.

TABLE \ref{tab:Effect_Diff_Scale} shows the performance comparison between the multi-scale and single-scale models when it detects the flying bird object. It can be seen from the results that the detection accuracy of the single-scale model structure is even better than that of the multi-scale model structure ($\text{AP}_{50}$ is increased by 1.8\%). Meanwhile, the inference speed of the single-scale model structure is slightly faster than that of the multi-scale model structure. Considering comprehensively, the single-scale model structure is more suitable for detecting flying birds in surveillance videos.

\begin{table}[!ht]
\caption{Effect of multi-scale model structure and single-scale model structure on detection performance.\label{tab:Effect_Diff_Scale}}
\centering
\begin{tabular}{c|c|c c c c}
\hline
Model Structure  & Frames & $\text{AP}_{50}$ & $\text{AP}_{75}$  & AP & \makecell[c]{Inference\\Time (s)} \\
\hline \hline
Multi-scale  & 5  & 0.744 & 0.276 & 0.343 & 0.0072\\
Single-scale  & 5 & 0.762 & 0.371 & 0.395 & 0.0070\\
\hline
\end{tabular}
\end{table}

\subsubsection{Comparative Experiments of Different Label Assignment Strategies}\label{ablation_analysis_3}

Since the flying bird object is a non-regular object in the bounding box, adopting the static label assignment method is unsuitable. In this paper, we design detection performance comparison experiments for models trained under different label assignment strategies. Specifically, two label assignment methods, Shrinking Bounding Box and Center Gaussian, are designed according to Foveabox \cite{2020_Kong_FoveaBox} and CenterNet \cite{2019_Duan_CenterNet} to compare with the SimOTA-OC label assignment method in this paper. Except for the label assignment method, the input mode, model structure, and loss function are the same as in this paper.

TABLE \ref{tab:Effect_Diff_Label_Assignment} shows the detection performance comparison results of the flying bird object detection models trained by the three label assignment strategies. As can be seen from the results, the detection accuracy $\text{AP}_{50}$ of the Center Gaussian label assignment method is 3.4\% higher than that of the Shrinking Bounding Box label assignment method, which is because the Center Gaussian gives lower weight to the anchor point samples close to the boundary of the bounding box compared with the Shrinking Bounding Box label assignment method, thereby weakening the impact of inaccurate sample assignment. The detection accuracy $\text{AP}_{50}$ of the proposed SimOTA-OC label assignment method is 1.2\% higher than that of the Center Gaussian label assignment method, which proves that the dynamic label assignment strategy is superior to the static label assignment strategy for the flying birds in the surveillance video. 

\begin{table}[!ht]
\caption{Effect of different label assignment method on detection performance.\label{tab:Effect_Diff_Label_Assignment}}
\centering
\begin{tabular}{c|c|c c c}
\hline
Label Assignment Method  & Frames & $\text{AP}_{50}$ & $\text{AP}_{75}$  & AP \\
\hline \hline
Shrinking Bounding Box  & 5  & 0.716 & 0.276 & 0.349 \\
Center Gaussian  & 5 & 0.750 & 0.370 & 0.393 \\
SimOTA-OC & 5 & 0.762 & 0.371 & 0.395 \\
\hline
\end{tabular}
\end{table}

\subsection{Compared with Other Methods}\label{comparative_analysis}
 
To verify the advancement of the proposed FBOD-SV, We selected the current advanced object detection methods (such as YOLOV5l \cite{yolov5_2021}, YOLOV6l \cite{2022_Chuyi_yolov6}, YOLOV7l \cite{2023_Wang_yolov7}, YOLOV8l \cite{2023_Guang_yolov8}, YOLOXl \cite{2021_Zheng_YOLOX}, SSD \cite{2016_Liu_SSD}, Foveabox \cite{2020_Kong_FoveaBox}, CenterNet) \cite{2019_Duan_CenterNet}), video object detection methods (FGFA \cite{2017_Zhu_FGFA}, SELSA \cite{2019_Wu_SELSA}, Temporal RoI Align \cite{2021_Tao_Temporal_RoI_Align}) and our previous work FBOD-BMI \cite{2024_sun_Flying_Bird} conducted quantitative and qualitative comparison experiments with FBOD-SV. These methods use the relevant open-source code. Among them, SSD \cite{2016_Liu_SSD}, Foveabox \cite{2020_Kong_FoveaBox}, CenterNet \cite{2019_Duan_CenterNet} use the MMDetection open-source framework \cite{mmdetection}; FGFA \cite{2017_Zhu_FGFA}, SELSA \cite{2019_Wu_SELSA}, and Temporal RoI Align \cite{2021_Tao_Temporal_RoI_Align} uses MMTracking open-source framework \cite{mmtrack2020}. In the object detection algorithms, YOLOV5l \cite{yolov5_2021}, YOLOV7l \cite{2023_Wang_yolov7}, SSD \cite{2016_Liu_SSD}, Foveabox \cite{2020_Kong_FoveaBox}, CenterNet \cite{2019_Duan_CenterNet} adopt the strategy of static label allocation, and YOLOV6l \cite{2022_Chuyi_yolov6}, YOLOXl \cite{2021_Zheng_YOLOX}, YOLOV8l \cite{2023_Guang_yolov8} adopt the strategy of dynamic label allocation. Object detection algorithms use the information of a single frame image to detect objects, while video object detection methods use the information of multiple frames of images. In addition, when training the model in FBOD-BMI \cite{2024_sun_Flying_Bird}, augmentation operations such as random Cropping, random Center Flipping, and random HSV are added (All data augmentations perform the same operation on $n$ consecutive frames).

The quantitative comparison experimental results are shown in TABLE \ref{tab:compare}. The $\text{AP}_{50}$ of the proposed FBOD-SV reaches 76.2\%, which is 24.9\% higher than that of SSD \cite{2016_Liu_SSD}, 22.6\% higher than that of YOLOV8l \cite{2023_Guang_yolov8}, 33.4\% higher than that of SELSA \cite{2019_Wu_SELSA}, and 3.1\% higher than FBOD-BMI \cite{2024_sun_Flying_Bird}. At the same time, the detection speed of the proposed method is twice that of FBOD-BMI, reaching 59.87fps to meet the real-time requirement of the bird object detection task in surveillance video.

SSD \cite{2016_Liu_SSD} not only uses single-frame image as input, but its label allocation strategy is also static. However, the flying bird object in the surveillance video is not regular in the bounding box, so the static label allocation strategy will have more wrong assignments, affecting the algorithm's accuracy. YOLOV8l \cite{2023_Guang_yolov8} adopts the Task Aligned Assigner dynamic label assignment strategy and excellent model structure, so its detection performance is stronger than other object detection methods. However, it also uses single-frame image as input, and the effect is still not ideal when applied to the flying birds in the surveillance video because most of the single-frame image features of the flying birds in the surveillance video are not obvious. Video object detection methods use the information from multiple frames of images. However, they extract the information of a single frame of image in the initial feature extraction, which will lead to the loss of the bird object with unobvious features in the process of feature extraction and the formation of wrong features in the feature aggregation, resulting in the decline of detection accuracy. FBOD-BMI \cite{2024_sun_Flying_Bird} adopts a two-stage method to make full use of the motion information of the flying bird object to detect the flying bird object. At the same time, ConvLSTM is used at the input of the model to aggregate the spatio-temporal information of adjacent multiple frames of the flying bird object, enhance the characteristics of the flying bird object, and improve the detection rate of the flying bird object. Therefore, FBOD-BMI \cite{2024_sun_Flying_Bird} has achieved good results. Compared with FBOD-BMI \cite{2024_sun_Flying_Bird}, the proposed FBOD-SV uses Co-Attention-FA to aggregate the spatio-temporal information of flying bird objects. At the same time, a detection model more suitable for the flying bird object in the surveillance video is designed, the SimOTA-OC dynamic label allocation strategy is adopted, and the $\text{AP}_{50}$ of the proposed FBOD method is increased by 3.1\%.

\begin{table*}[!ht]
\caption{Comparison with other object detection methods. The $\text{AP}_\text{S}$ represents the AP that only computes small size ($\textless 32 \times 32$  pixels) flying bird objects. $\text{AP}_\text{M}$ ($\geq 32 \times 32$ and $\leq 96 \times 96$  pixels) and $\text{AP}_\text{L}$ ($\textgreater 96 \times 96$  pixels) are similar to $\text{AP}_\text{S}$.\label{tab:compare}}
\centering
\begin{threeparttable}
\begin{tabular}{c|c|c| c c c c c c c}
\hline
Method  & Backbnoe & Image Size  & $\text{AP}_{50}$ & $\text{AP}_{75}$  & AP & $\text{AP}_\text{S}$ & $\text{AP}_\text{M}$ & $\text{AP}_\text{L}$ & \makecell[c]{Detection\\Speed (fps)}\\
\hline \hline
YOLOV5l \cite{yolov5_2021} & CSPResNet50 & 640$\times$640 & 0.503 & 0.312 & 0.291 & 0.093 & 0.465 & 0.437 & 83.68\\
YOLOV6l \cite{2022_Chuyi_yolov6} & EfficientRep & 640$\times$640 & 0.519 & 0.320 & 0.219 & 0.079 & 0.481 & 0.643 & 65.77\\
YOLOV7l\tnote{1} \cite{2023_Wang_yolov7} & CSPDarkNet53 & 640$\times$640 & 0.423 & 0.253 & 0.239 & 0.055 & 0.416 & 0.378 & 78.61\\
YOLOV8l \cite{2023_Guang_yolov8} & CSPResNet50 & 640$\times$640 & 0.536 & 0.345 & 0.322 & 0.085 & 0.479 & 0.634 & 67.52\\
YOLOXl \cite{2021_Zheng_YOLOX} & CSPResNet50 & 640$\times$640 & 0.528 & 0.349 & 0.306 & 0.089 & 0.491 & 0.596 & 48.29\\
SSD \cite{2016_Liu_SSD} & SSDVGG16 & 640$\times$640 & 0.513 & 0.283 & 0.275 & 0.073 & 0.449 & 0.496 & 59.12\\
Foveabox \cite{2020_Kong_FoveaBox} & ResNet101 & 640$\times$640 & 0.434 & 0.128 & 0.188 & 0.033 & 0.338 & 0.418 & 42.82\\
CenterNet \cite{2019_Duan_CenterNet} & ResNet18 & 640$\times$640 & 0.469 & 0.281 & 0.256 & 0.056 & 0.428 & 0.519 & 38.44\\
FGFA \cite{2017_Zhu_FGFA} & ResNet101 & 1000$\times$600 & 0.281 & 0.212 & 0.175 & 0.001 & 0.338 & 0.185 & 17.65\\
SELSA \cite{2019_Wu_SELSA} & ResNet101 & 1000$\times$600 & 0.428 & 0.290 & 0.248 & 0.017 & 0.471 & 0.565 & 14.88\\
Temporal RoI Align \cite{2021_Tao_Temporal_RoI_Align} & ResNet101 & 1000$\times$600 & 0.424 & 0.279 & 0.247 & 0.018 & 0.469 & 0.519 & 7.28\\
FBOD-BMI\tnote{2} \cite{2024_sun_Flying_Bird} & CSPDarkNet53 & 672$\times$384 & 0.731 & 0.323 & 0.356 & 0.186 & 0.487 & 0.266 & 28.56\\
FBOD-SV (this paper) & CSPDarkNet53 & 672$\times$384 & 0.762 & 0.371 & 0.395 & 0.203 & 0.545 & 0.659 & 59.87\\
\hline
\end{tabular}
\begin{tablenotes}
        \footnotesize
        \item[1] We have trained YOLOV7l for multiple rounds using our dataset, and the results are still poor. Maybe it's because we didn't tune the parameters properly.
        \item[2] After adding data augmentation (especially random cropping), the detection performance of FBOD-BMI is improved.
\end{tablenotes}
\end{threeparttable}
\end{table*}

\begin{figure*}[!htp]
    \centering
    \subfloat[Situation 1]{
        \begin{minipage}[t]{0.225\linewidth}
        \centering
        \includegraphics[width=1\linewidth]{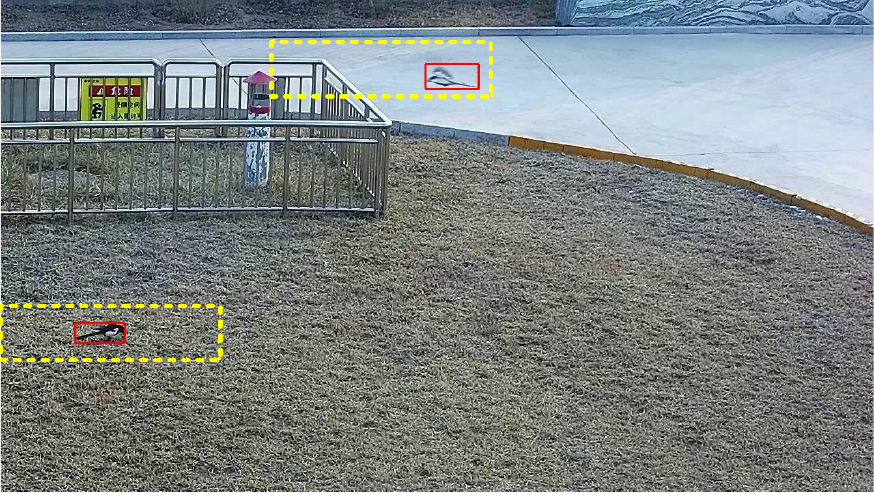}
        \label{S1}
        \end{minipage}
        }
    \subfloat[Situation 2]{
        \begin{minipage}[t]{0.225\linewidth}
        \centering
        \includegraphics[width=1\linewidth]{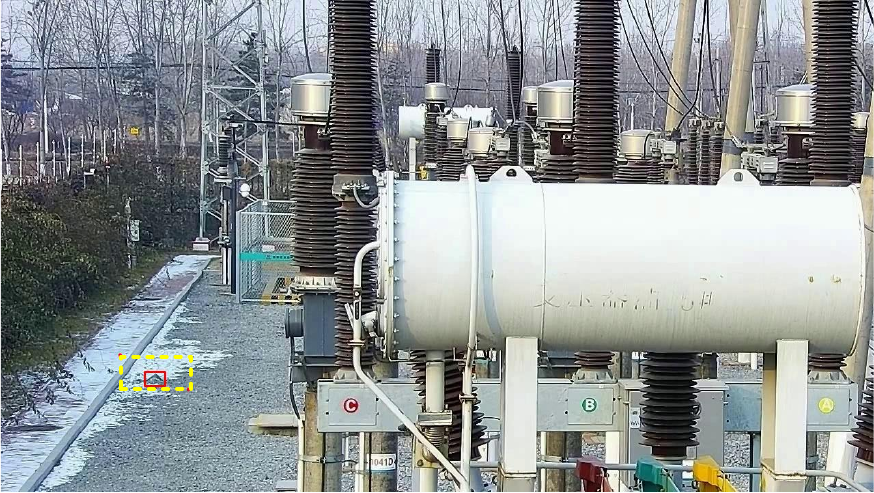}
        \label{S2}
        \end{minipage}
        }
    \subfloat[Situation 3]{
        \begin{minipage}[t]{0.225\linewidth}
        \centering
        \includegraphics[width=1\linewidth]{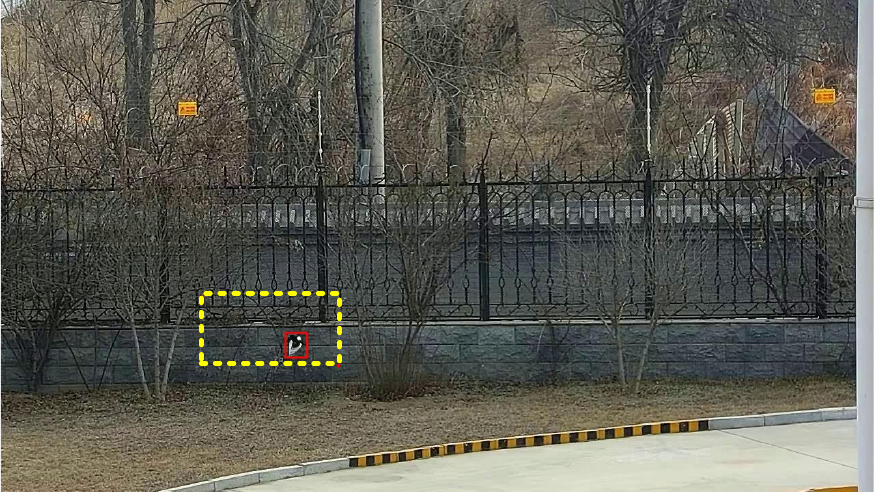}
        \label{S3}
        \end{minipage}
        }
    \subfloat[Situation 4]{
        \begin{minipage}[t]{0.225\linewidth}
        \centering
        \includegraphics[width=1\linewidth]{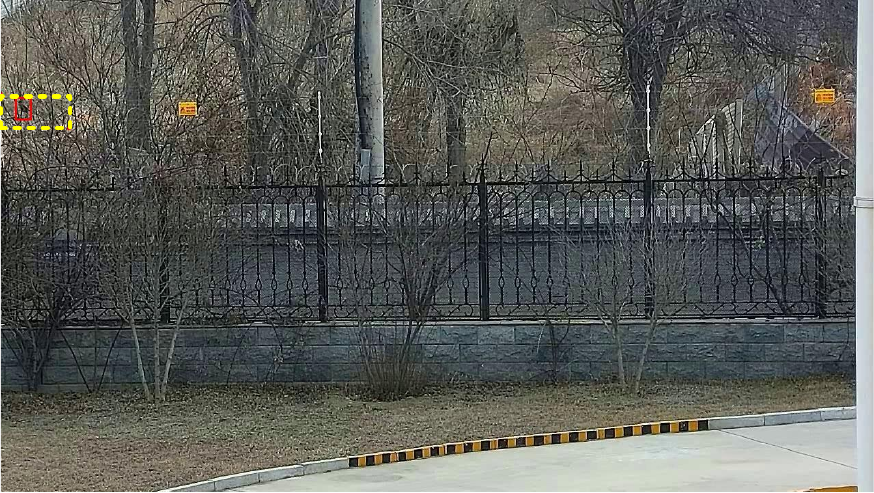}
        \label{S4}
        \end{minipage}
        }
    \caption{Four typical Situations.}
    \label{Situation}
\end{figure*}

In qualitative experiments, we selected three representative methods (YOLOV8l \cite{2023_Guang_yolov8}, SELSA \cite{2019_Wu_SELSA}, and FBOD-BMI \cite{2024_sun_Flying_Bird}) for experiments and selected four typical Situations to compare and analyze the experimental results. The four Situations are shown in Fig. \ref{Situation}. In Situation 1, there are two flying bird objects with clear appearance features. In Situation 2, the characteristics of the flying bird object are not obvious in the single-frame image, but the movement is obvious in the continuous image frames. In Situation 3, the appearance shape of the flying bird object changes greatly during its flight. In Situation 4, the characteristics of the single frame image of the flying bird object are not obvious, and the movement on the continuous image frames is not obvious either. 

\begin{figure*}[!htp]
    \centering
    \subfloat[Situation 1]{
        \begin{minipage}[t]{0.45\linewidth}
        \centering
        \includegraphics[width=1\linewidth]{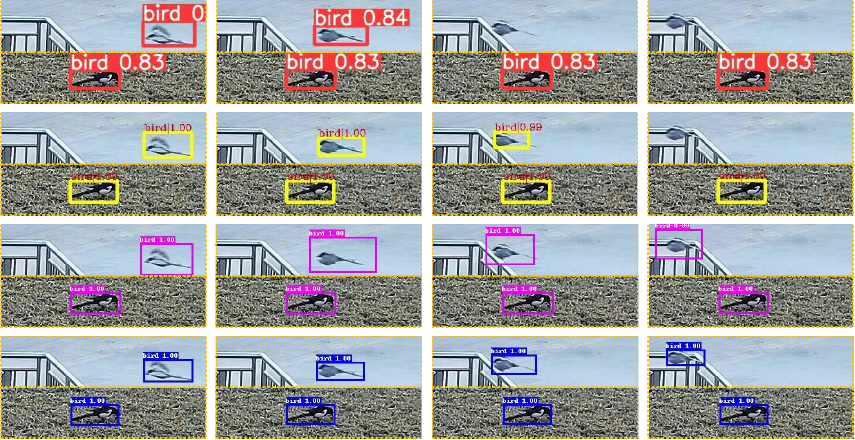}
        \label{s1_result}
        \scriptsize{$1^\text{{st}}$ frame~~~~~~~~~~~~$2^\text{{nd}}$ frame~~~~~~~~~~~~$3^\text{{rd}}$ frame~~~~~~~~~~~~$4^\text{{th}}$ frame}
        \end{minipage}
        }\hspace{3mm}
    \subfloat[Situation 2]{
        \begin{minipage}[t]{0.45\linewidth}
        \centering
        \includegraphics[width=1\linewidth]{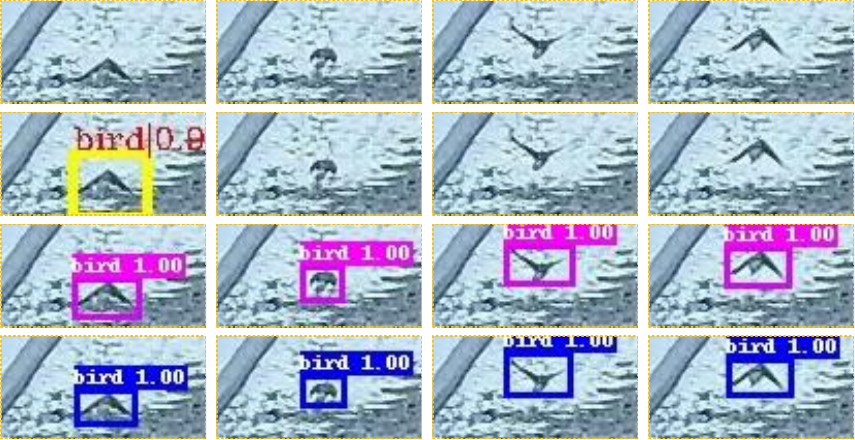}
        \label{s2_result}
        \scriptsize{$1^\text{{st}}$ frame~~~~~~~~~~~~$2^\text{{nd}}$ frame~~~~~~~~~~~~$3^\text{{rd}}$ frame~~~~~~~~~~~~$4^\text{{th}}$ frame}
        \end{minipage}
        }
    
    \vspace{-1mm}
    
    \subfloat[Situation 3]{
        \begin{minipage}[t]{0.45\linewidth}
        \centering
        \includegraphics[width=1\linewidth]{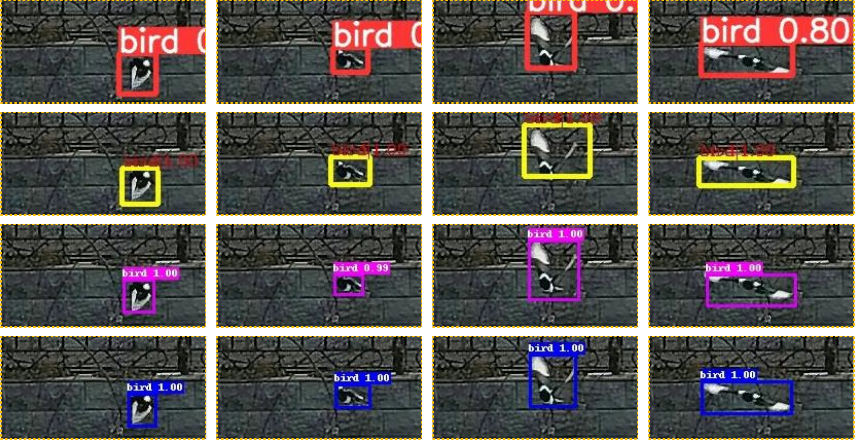}
        \label{s3_result}
        \scriptsize{$1^\text{{st}}$ frame~~~~~~~~~~~~$2^\text{{nd}}$ frame~~~~~~~~~~~~$3^\text{{rd}}$ frame~~~~~~~~~~~~$4^\text{{th}}$ frame}
        \end{minipage}
        }\hspace{3mm}
    \subfloat[Situation 4]{
        \begin{minipage}[t]{0.45\linewidth}
        \centering
        \includegraphics[width=1\linewidth]{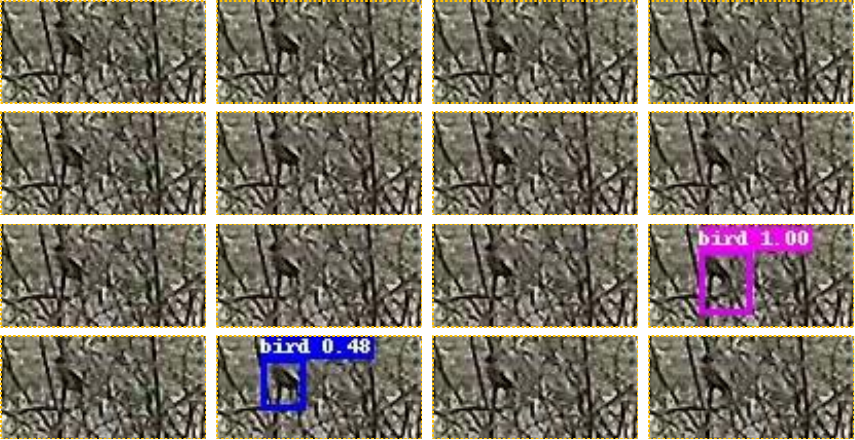}
        \label{s4_result}
        \scriptsize{$1^\text{{st}}$ frame~~~~~~~~~~~~$2^\text{{nd}}$ frame~~~~~~~~~~~~$3^\text{{rd}}$ frame~~~~~~~~~~~~$4^\text{{th}}$ frame}
        \end{minipage}
        }
    \caption{Screenshots of detection results on four consecutive frames by different methods in four typical Situations. The first row shows YOLOV8l detection results, the second row shows SELSA detection results, the third row shows FBOD-BMI detection results, and the fourth row shows FBOD-SV detection results.}
    \label{result}
\end{figure*}

The experimental results are shown in Fig. \ref{result}. From the experimental results, it can be seen that in Situation 1, all methods can detect the flying bird objects (when the bird object is slightly similar to the background, YOLOV8l \cite{2023_Guang_yolov8} and SELSA \cite{2019_Wu_SELSA} will miss detection in the third and fourth frames). In Situation 2, YOLOV8l \cite{2023_Guang_yolov8} and SELSA \cite{2019_Wu_SELSA} basically cannot detect the flying bird object, while FBOD-BMI \cite{2024_sun_Flying_Bird} and FBOD-SV can detect the flying bird object well, which confirms the analysis in the quantitative experiment: that is, YOLOV8l \cite{2023_Guang_yolov8} and SELSA \cite{2019_Wu_SELSA} cannot effectively extract the features of flying bird objects when most single-frame image features of flying bird objects are not obvious. In Situation 3, the detection box of FBOD-BMI \cite{2024_sun_Flying_Bird} is not as accurate as that of YOLOV8l \cite{2023_Guang_yolov8} and FBOD-SV (the detection box of FBOD-BMI \cite{2024_sun_Flying_Bird} in Situation 1 is not as accurate as that of FBOD-SV either), which indicates that the effect of static label allocation strategy is worse than that of dynamic label allocation strategy. In Situation 4, all methods do not achieve good results. The reason for YOLOV8l \cite{2023_Guang_yolov8} and SELSA \cite{2019_Wu_SELSA} detection failure is the same as Situation 1. FBOD-BMI \cite{2024_sun_Flying_Bird} and FBOD-SV aggregate the features of the flying bird object on consecutive frames of images. When the image features of the flying bird object are not obvious on consecutive frames, the detection will fail (In the subsequent research, we are considering using the global information of the video to solve this problem).

\section{Conclusion}\label{Conclusion}

In this paper, we have conducted an in-depth study of flying bird objects in surveillance videos. It is found that the characteristics of the flying bird object in the surveillance video are not obvious in a single-frame image, and the flying bird object is small and asymmetric in most cases. According to these characteristics, a method, FBOD-SV, for detecting flying birds in surveillance video was proposed. Firstly, addressing the issue of non-obvious features in single frames of flying bird objects in surveillance video, a new feature aggregation module, namely the Co-Attention-FA module, was devised. This module aggregates the features of the flying bird object by exploiting the correlation across multiple consecutive frames of images. Secondly, considering the multi-scale (particularly small size in most cases) character of flying birds in surveillance videos, the FBOD-Net was designed with a structure that first down-samples and then up-samples. A large feature layer that combines fine spatial information and large receptive field information is used to detect the flying birds. Then, considering the asymmetric and irregular characteristics of flying birds in surveillance video, we simplified the SimOTA dynamic label allocation method and proposed the SimOTA-OC dynamic label allocation strategy for One-Category object detection. This strategy aims to address the issue of label allocation posed by the irregular flying patterns of birds. Finally, a series of quantitative and qualitative experiments were designed to prove the effectiveness of the proposed bird object detection method in surveillance video, and the following conclusions are drawn:
\begin{enumerate}
\item{The features of the single-frame image of the flying bird object in the surveillance video are not obvious, and their features need to be aggregated before input into the model. The Co-Attention-FA module proposed in this paper can effectively aggregate the flying bird object's features on consecutive frames.}
\item{In most cases, the flying birds in surveillance videos are small-scale objects. Using a large feature layer that combines fine spatial information and large receptive field information to detect special multi-scale flying birds can achieve better detection performance.}
\item{In most cases, the flying birds in the surveillance video are asymmetric. In the process of label allocation, it is advisable to use a dynamic label allocation method.}
\end{enumerate}


%





\ifCLASSOPTIONcaptionsoff
  \newpage
\fi





\bibliographystyle{IEEEtran}
\bibliography{IEEEabrv,Bibliography}

\begin{thebibliography}{10}
\providecommand{\url}[1]{#1}
\csname url@rmstyle\endcsname
\providecommand{\newblock}{\relax}
\providecommand{\bibinfo}[2]{#2}
\providecommand\BIBentrySTDinterwordspacing{\spaceskip=0pt\relax}
\providecommand\BIBentryALTinterwordstretchfactor{4}
\providecommand\BIBentryALTinterwordspacing{\spaceskip=\fontdimen2\font plus
\BIBentryALTinterwordstretchfactor\fontdimen3\font minus \fontdimen4\font\relax}
\providecommand\BIBforeignlanguage[2]{{%
\expandafter\ifx\csname l@#1\endcsname\relax
\typeout{** WARNING: IEEEtran.bst: No hyphenation pattern has been}%
\typeout{** loaded for the language `#1'. Using the pattern for}%
\typeout{** the default language instead.}%
\else
\language=\csname l@#1\endcsname
\fi
#2}}

\bibitem{2016_Hoffmann_multistatic_radar}
F.~Hoffmann, M.~Ritchie, F.~Fioranelli, A.~Charlish, and H.~Griffiths, ``Micro-doppler based detection and tracking of uavs with multistatic radar,'' in \emph{2016 IEEE Radar Conference (RadarConf)}, 2016, pp. 1--6.

\bibitem{2017_Jahangirstaring_radar}
M.~Jahangir, C.~J. Baker, and G.~A. Oswald, ``Doppler characteristics of micro-drones with l-band multibeam staring radar,'' in \emph{2017 IEEE Radar Conference (RadarConf)}, 2017, pp. 1052--1057.

\bibitem{2022_Ye_CT-Net}
T.~Ye, J.~Zhang, Y.~Li, X.~Zhang, Z.~Zhao, and Z.~Li, ``Ct-net: An efficient network for low-altitude object detection based on convolution and transformer,'' \emph{IEEE Transactions on Instrumentation and Measurement}, vol.~71, pp. 1--12, 2022.

\bibitem{2022_Zheng_Foreign_Objects}
S.~Zheng, Z.~Wu, Y.~Xu, and Z.~Wei, ``Intrusion detection of foreign objects in overhead power system for preventive maintenance in high-speed railway catenary inspection,'' \emph{IEEE Transactions on Instrumentation and Measurement}, vol.~71, pp. 1--12, 2022.

\bibitem{2023_Ni_Improved_SSD}
J.~Ni, K.~Shen, Y.~Chen, and S.~X. Yang, ``An improved ssd-like deep network-based object detection method for indoor scenes,'' \emph{IEEE Transactions on Instrumentation and Measurement}, vol.~72, pp. 1--15, 2023.

\bibitem{2023_Ye_Real-Time_Object_Detection_Network_in_UAV-Vision}
T.~Ye, W.~Qin, Z.~Zhao, X.~Gao, X.~Deng, and Y.~Ouyang, ``Real-time object detection network in uav-vision based on cnn and transformer,'' \emph{IEEE Transactions on Instrumentation and Measurement}, vol.~72, pp. 1--13, 2023.

\bibitem{2014_Girshick_RCNN}
R.~Girshick, J.~Donahue, T.~Darrell, and J.~Malik, ``Rich feature hierarchies for accurate object detection and semantic segmentation,'' in \emph{2014 IEEE Conference on Computer Vision and Pattern Recognition}, 2014, pp. 580--587.

\bibitem{2015_Girshick_Fast_RCNN}
R.~Girshick, ``Fast r-cnn,'' in \emph{2015 IEEE International Conference on Computer Vision (ICCV)}, 2015, pp. 1440--1448.

\bibitem{2017_Ren_Faster_RCNN}
S.~Ren, K.~He, R.~Girshick, and J.~Sun, ``Faster r-cnn: Towards real-time object detection with region proposal networks,'' \emph{IEEE Transactions on Pattern Analysis and Machine Intelligence}, vol.~39, no.~6, pp. 1137--1149, 2017.

\bibitem{2016_Liu_SSD}
W.~Liu, D.~Anguelov, D.~Erhan, C.~Szegedy, S.~Reed, C.~Y. Fu, and A.~C. Berg, ``Ssd: Single shot multibox detector,'' in \emph{2016 European Conference on Computer Vision (ECCV)}, 2016.

\bibitem{2016_Redmon_YOLO}
J.~Redmon, S.~Divvala, R.~Girshick, and A.~Farhadi, ``You only look once: Unified, real-time object detection,'' in \emph{2016 IEEE Conference on Computer Vision and Pattern Recognition (CVPR)}, 2016, pp. 779--788.

\bibitem{2017_Redmon_YOLOV2}
J.~Redmon and A.~Farhadi, ``Yolo9000: Better, faster, stronger,'' in \emph{2017 IEEE Conference on Computer Vision and Pattern Recognition (CVPR)}, 2017, pp. 6517--6525.

\bibitem{2018_Redmon_YOLOv3}
J.~\vspace{0mm} Redmon and A.~Farhadi, ``Yolov3: An incremental improvement,'' \emph{arXiv e-prints}, 2018.

\bibitem{2020_Bochkovskiy_YOLOv4}
A.~Bochkovskiy, C.~Y. Wang, and H.~Liao, ``Yolov4: Optimal speed and accuracy of object detection,'' 2020.

\bibitem{yolov5_2021}
Y.~Contributors, ``You only look once version 5,'' \url{https://github.com/ultralytics/yolov5}, 2021.

\bibitem{2021_Zheng_YOLOX}
\BIBentryALTinterwordspacing
Z.~Ge, S.~Liu, F.~Wang, Z.~Li, and J.~Sun, ``{YOLOX:} exceeding {YOLO} series in 2021,'' \emph{CoRR}, vol. abs/2107.08430, 2021. [Online]. Available: \url{https://arxiv.org/abs/2107.08430}
\BIBentrySTDinterwordspacing

\bibitem{2022_Chuyi_yolov6}
C.~Li, L.~Li, H.~Jiang, K.~Weng, Y.~Geng, L.~Li, Z.~Ke, Q.~Li, M.~Cheng, W.~Nie, Y.~Li, B.~Zhang, Y.~Liang, L.~Zhou, X.~Xu, X.~Chu, X.~Wei, and X.~Wei, ``Yolov6: A single-stage object detection framework for industrial applications,'' 2022.

\bibitem{2023_Wang_yolov7}
\BIBentryALTinterwordspacing
C.-Y. Wang, A.~Bochkovskiy, and H.-Y.~M. Liao, ``Yolov7: Trainable bag-of-freebies sets new state-of-the-art for real-time object detectors,'' \emph{2023 IEEE/CVF Conference on Computer Vision and Pattern Recognition (CVPR)}, Jun 2023. [Online]. Available: \url{http://dx.doi.org/10.1109/CVPR52729.2023.00721}
\BIBentrySTDinterwordspacing

\bibitem{2023_Guang_yolov8}
G.~J.~N. Ang, A.~K. Goil, H.~Chan, J.~J. Lew, X.~C. Lee, R.~B.~A. Mustaffa, T.~Jason, Z.~T. Woon, and B.~Shen, ``A novel application for real-time arrhythmia detection using yolov8,'' 2023.

\bibitem{2015_ImageNet}
\BIBentryALTinterwordspacing
O.~Russakovsky, J.~Deng, H.~Su, J.~Krause, S.~Satheesh, S.~Ma, Z.~Huang, A.~Karpathy, A.~Khosla, M.~Bernstein, A.~C. Berg, and L.~Fei-Fei, ``Imagenet large scale visual recognition challenge,'' \emph{International Journal of Computer Vision}, vol. 115, no.~3, pp. 211--252, Dec 2015. [Online]. Available: \url{https://doi.org/10.1007/s11263-015-0816-y}
\BIBentrySTDinterwordspacing

\bibitem{2014_COCO}
T.-Y. Lin, M.~Maire, S.~Belongie, J.~Hays, P.~Perona, D.~Ramanan, P.~Doll{\'a}r, and C.~L. Zitnick, ``Microsoft coco: Common objects in context,'' in \emph{Computer Vision -- ECCV 2014}, D.~Fleet, T.~Pajdla, B.~Schiele, and T.~Tuytelaars, Eds.\hskip 1em plus 0.5em minus 0.4em\relax Cham: Springer International Publishing, 2014, pp. 740--755.

\bibitem{2010_Pascal_VOC}
\BIBentryALTinterwordspacing
M.~Everingham, L.~Van~Gool, C.~K.~I. Williams, J.~Winn, and A.~Zisserman, ``The pascal visual object classes (voc) challenge,'' \emph{International Journal of Computer Vision}, vol.~88, no.~2, pp. 303--338, Jun 2010. [Online]. Available: \url{https://doi.org/10.1007/s11263-009-0275-4}
\BIBentrySTDinterwordspacing

\bibitem{2018_Tianhuang_skeleton_flying_bird}
\BIBentryALTinterwordspacing
T.~WU, X.~LUO, and Q.~XU, ``A new skeleton based flying bird detection method for low-altitude air traffic management,'' \emph{Chinese Journal of Aeronautics}, vol.~31, no.~11, pp. 2149--2164, 2018. [Online]. Available: \url{https://www.sciencedirect.com/science/article/pii/S1000936118300360}
\BIBentrySTDinterwordspacing

\bibitem{2011_Barnich_ViBe}
O.~Barnich and M.~Van~Droogenbroeck, ``Vibe: A universal background subtraction algorithm for video sequences,'' \emph{IEEE Transactions on Image Processing}, vol.~20, no.~6, pp. 1709--1724, 2011.

\bibitem{2019_Tian_GSD}
S.~Tian, X.~Cao, Y.~Li, X.~Zhen, and B.~Zhang, ``Glance and stare: Trapping flying birds in aerial videos by adaptive deep spatio-temporal features,'' \emph{IEEE Transactions on Circuits and Systems for Video Technology}, vol.~29, no.~9, pp. 2748--2759, 2019.

\bibitem{2024_sun_Flying_Bird}
Z.-W. Sun, Z.-X. Hua, H.-C. Li, and H.-Y. Zhong, ``Flying bird object detection algorithm in surveillance video based on motion information,'' \emph{IEEE Transactions on Instrumentation and Measurement}, vol.~73, pp. 1--15, 2024.

\bibitem{2017_Zhu_DFF}
X.~Zhu, Y.~Xiong, J.~Dai, L.~Yuan, and Y.~Wei, ``Deep feature flow for video recognition,'' in \emph{2017 IEEE Conference on Computer Vision and Pattern Recognition (CVPR)}, 2017, pp. 4141--4150.

\bibitem{2017_Zhu_FGFA}
X.~Zhu, Y.~Wang, J.~Dai, L.~Yuan, and Y.~Wei, ``Flow-guided feature aggregation for video object detection,'' in \emph{2017 IEEE International Conference on Computer Vision (ICCV)}, 2017, pp. 408--417.

\bibitem{2018_Zhu_Towards_High_Performance}
X.~Zhu, J.~Dai, L.~Yuan, and Y.~Wei, ``Towards high performance video object detection,'' in \emph{2018 IEEE/CVF Conference on Computer Vision and Pattern Recognition}, 2018, pp. 7210--7218.

\bibitem{2017_Hetang_Impression_Network}
C.~Hetang, H.~Qin, S.~Liu, and J.~Yan, ``Impression network for video object detection,'' \emph{arXiv}, 2017.

\bibitem{2019_Wu_SELSA}
H.~Wu, Y.~Chen, N.~Wang, and Z.-X. Zhang, ``Sequence level semantics aggregation for video object detection,'' in \emph{2019 IEEE/CVF International Conference on Computer Vision (ICCV)}, 2019, pp. 9216--9224.

\bibitem{2021_Tao_Temporal_RoI_Align}
T.~Gong, K.~Chen, X.~Wang, Q.~Chu, F.~Zhu, D.~Lin, N.~Yu, and H.~Feng, ``Temporal roi align for video object recognition,'' in \emph{The Thirty-Fifth AAAI Conference on Artificial Intelligence (AAAI-21)}, 2021, pp. 1442--1450.

\bibitem{2017_Lin_Feature_Pyramid_Networks}
T.-Y. Lin, P.~Dollár, R.~Girshick, K.~He, B.~Hariharan, and S.~Belongie, ``Feature pyramid networks for object detection,'' in \emph{2017 IEEE Conference on Computer Vision and Pattern Recognition (CVPR)}, 2017, pp. 936--944.

\bibitem{2018_Liu_Path_Aggregation_Network}
S.~Liu, L.~Qi, H.~Qin, J.~Shi, and J.~Jia, ``Path aggregation network for instance segmentation,'' in \emph{2018 IEEE/CVF Conference on Computer Vision and Pattern Recognition}, 2018, pp. 8759--8768.

\bibitem{2019_Tian_FCOS}
Z.~Tian, C.~Shen, H.~Chen, and T.~He, ``Fcos: Fully convolutional one-stage object detection,'' in \emph{2019 IEEE/CVF International Conference on Computer Vision (ICCV)}, 2019, pp. 9626--9635.

\bibitem{2020_Kong_FoveaBox}
T.~Kong, F.~Sun, H.~Liu, Y.~Jiang, L.~Li, and J.~Shi, ``Foveabox: Beyound anchor-based object detection,'' \emph{IEEE Transactions on Image Processing}, vol.~29, pp. 7389--7398, 2020.

\bibitem{2020_Li_Learning_From_Noisy_Anchors}
H.~Li, Z.~Wu, C.~Zhu, C.~Xiong, R.~Socher, and L.~S. Davis, ``Learning from noisy anchors for one-stage object detection,'' in \emph{2020 IEEE/CVF Conference on Computer Vision and Pattern Recognition (CVPR)}, 2020, pp. 10\,585--10\,594.

\bibitem{2020_Kim_PAA}
\BIBentryALTinterwordspacing
K.~Kim and H.~S. Lee, ``Probabilistic anchor assignment with iou prediction for object detection,'' \emph{Lecture Notes in Computer Science}, p. 355–371, 2020. [Online]. Available: \url{http://dx.doi.org/10.1007/978-3-030-58595-2_22}
\BIBentrySTDinterwordspacing

\bibitem{2020_Benjin_AutoAssign}
B.~Zhu, J.~Wang, Z.~Jiang, F.~Zong, S.~Liu, Z.~Li, and J.~Sun, ``Autoassign: Differentiable label assignment for dense object detection,'' 2020.

\bibitem{2021_Ge_OTA}
Z.~Ge, S.~Liu, Z.~Li, O.~Yoshie, and J.~Sun, ``Ota: Optimal transport assignment for object detection,'' in \emph{2021 IEEE/CVF Conference on Computer Vision and Pattern Recognition (CVPR)}, 2021, pp. 303--312.

\bibitem{2015_JaderBerg_STN}
M.~Jaderberg, K.~Simonyan, A.~Zisserman, and K.~Kavukcuoglu, ``\BIBforeignlanguage{English}{Spatial transformer networks},'' vol. 2015-January, 2015, pp. 2017 -- 2025.

\bibitem{2020_Wang_CSPNet}
C.-Y. Wang, H.-Y. Mark~Liao, Y.-H. Wu, P.-Y. Chen, J.-W. Hsieh, and I.-H. Yeh, ``Cspnet: A new backbone that can enhance learning capability of cnn,'' in \emph{2020 IEEE/CVF Conference on Computer Vision and Pattern Recognition Workshops (CVPRW)}, 2020, pp. 1571--1580.

\bibitem{2019_Duan_CenterNet}
K.~Duan, S.~Bai, L.~Xie, H.~Qi, Q.~Huang, and Q.~Tian, ``Centernet: Keypoint triplets for object detection,'' in \emph{2019 IEEE/CVF International Conference on Computer Vision (ICCV)}, 2019, pp. 6568--6577.

\bibitem{2021_zheng_CIOU}
Z.~Zheng, P.~Wang, D.~Ren, W.~Liu, R.~Ye, Q.~Hu, and W.~Zuo, ``Enhancing geometric factors in model learning and inference for object detection and instance segmentation,'' \emph{IEEE Transactions on cybernetics}, vol.~52, no.~8, pp. 8574--8586, 2021.

\bibitem{mmdetection}
K.~Chen, J.~Wang, J.~Pang, Y.~Cao, Y.~Xiong, X.~Li, S.~Sun, W.~Feng, Z.~Liu, J.~Xu, Z.~Zhang, D.~Cheng, C.~Zhu, T.~Cheng, Q.~Zhao, B.~Li, X.~Lu, R.~Zhu, Y.~Wu, J.~Dai, J.~Wang, J.~Shi, W.~Ouyang, C.~C. Loy, and D.~Lin, ``{MMDetection}: Open mmlab detection toolbox and benchmark,'' \emph{arXiv preprint arXiv:1906.07155}, 2019.

\bibitem{mmtrack2020}
M.~Contributors, ``{MMTracking: OpenMMLab} video perception toolbox and benchmark,'' \url{https://github.com/open-mmlab/mmtracking}, 2020.

\end{thebibliography}

\vfill


\end{document}